\documentclass{article}
\usepackage{amsmath}

\usepackage[preprint]{neurips_2025}

\usepackage[utf8]{inputenc} % allow utf-8 input
\usepackage[T1]{fontenc}    % use 8-bit T1 fonts
\usepackage{hyperref}       % hyperlinks
\usepackage{url}            % simple URL typesetting
\usepackage{booktabs}       % professional-quality tables
\usepackage{makecell}      % for \makecell in tables
\usepackage{amsfonts}       % blackboard math symbols
\usepackage{nicefrac}       % compact symbols for 1/2, etc.
\usepackage{microtype}      % microtypography
\usepackage{xcolor}         % colors
\usepackage{wasysym}
\usepackage{footnote}
\usepackage{multirow}
\usepackage{subfigure}
\usepackage{pdfpages}
\usepackage{tabularx}
\usepackage[font={small}]{caption}
\usepackage{amssymb}
\usepackage{tcolorbox}
\usepackage[lined,ruled,commentsnumbered]{algorithm2e}
\SetKwComment{Comment}{$\triangleright$\ }{}
\usepackage{threeparttable}
\usepackage{bbm}
\usepackage{subfigure}
\usepackage{wrapfig}
\usepackage{ragged2e}
\definecolor{citecol}{HTML}{2DDC0E}
\definecolor{tableofcontent}{HTML}{E63E15}
\definecolor{urlcol}{HTML}{2470D8}
\hypersetup{
    colorlinks=true,       % false: boxed links; true: colored links
    linkcolor=tableofcontent, 
    citecolor=citecol,        % color of links to bibliography
    urlcolor=black,           % color of external links
}

\title{GRExplainer: A Universal Explanation Method for Temporal Graph Neural Networks}

\author{
Xuyan Li\\
Xidian University\\
\texttt{xli77144@gmail.com}\\
\And
Jie Wang\\
Xidian University\\
\texttt{jwang1997@stu.xidian.edu.cn}
\And
Zheng Yan\\
Xidian University\\
\texttt{zyan@xidian.edu.cn}}

\begin{document}

\maketitle

\begin{abstract}
Dynamic graphs are widely used to represent evolving real-world networks. Temporal Graph Neural Networks (TGNNs) have emerged as a powerful tool for processing such graphs, but the lack of transparency and explainability limits their practical adoption. Research on TGNN explainability is still in its early stages and faces several key issues: (i) Current methods are tailored to specific TGNN types, restricting generality. (ii) They suffer from high computational costs, making them unsuitable for large-scale networks. (iii) They often overlook the structural connectivity of explanations and require prior knowledge, reducing user-friendliness. To address these issues, we propose GRExplainer, the first universal, efficient, and user-friendly explanation method for TGNNs. GRExplainer extracts node sequences as a unified feature representation, making it independent of specific input formats and thus applicable to both snapshot-based and event-based TGNNs (the major types of TGNNs). By utilizing breadth-first search and temporal information to construct input node sequences, GRExplainer reduces redundant computation and improves efficiency. To enhance user-friendliness, we design a generative model based on Recurrent Neural Networks (RNNs), enabling automated and continuous explanation generation. Experiments on six real-world datasets with three target TGNNs show that GRExplainer outperforms existing baseline methods in generality, efficiency, and user-friendliness.
\end{abstract}

\section{Introduction}
Dynamic graphs are well suited for modeling real-world networks with time-varying interactions, such as social, communication, and financial networks~\cite{min2021stgsn,li2024jointly,wang2021review}. To analyze such graphs, Temporal Graph Neural Networks (TGNNs) have been developed, which integrate both spatial and temporal features, unlike static GNNs that capture only spatial information within a fixed structure. Based on their dynamic graph representations, TGNNs can be categorized into snapshot-based (discrete-time)~\cite{pareja2020evolvegcn,hasanzadeh2019variational} and event-based (continuous-time)~\cite{xu2020inductive,rossi2020temporal} models: the former represents dynamic graphs as sequences of static snapshots observed at different timeslots, while the latter represents them as continuous streams of timestamped events. By effectively handling spatial-temporal dependencies, TGNNs have been widely applied in fraud detection~\cite{cheng2020graph}, social recommendation~\cite{song2019session}, network intrusion detection~\cite{khoury2024jbeil}, and other domains.

While TGNNs offer a powerful tool for dynamic graph analysis, their black-box nature prevents their usage in safety-critical areas~\cite{han2024rules}, highlighting the need for explainability. Explainability refers to the capability of a model to clarify its inner workings and decision-making processes in a way that is understandable to a specific audience~\cite{arrieta2020explainable}. It is significant for several reasons~\cite{yuan2022explainability,li2025can,wang2024trustguard}: (i) It improves the transparency of the model, thereby fostering user trust. (ii) It enables the detection of biases, unfair decisions, and spurious correlations, thus making the model easy to be accepted. (iii) It enhances model robustness by identifying potential vulnerabilities and providing insights into the design of resilient architectures~\cite{mankali2024insight}. Consequently, explainability is not merely a desirable property but a fundamental requirement for the safe, ethical, and trustworthy deployment of TGNNs.

Currently, there is a large body of research on explaining static GNNs~\cite{yuan2022explainability,li2025can,zhang2024trustworthy}, while the explainability of TGNNs remains underexplored. Notably, the explanation methods developed for static GNNs are not directly applicable to TGNNs due to the inherent complexity of dynamic graphs. First, dynamic graphs introduce a temporal dimension that plays a critical role in model predictions. For instance, there can be duplicate events occurring at different timestamps, which carry distinct meanings and varying degrees of importance. Static explanation methods fail to capture such temporal dependencies, resulting in sub-optimal explanations. Second, dynamic graphs can be represented in different ways, which complicates the design of universal explanation methods. As a result, explaining TGNNs is considerably more challenging than explaining static GNNs.

\begin{wraptable}{r}{0.5\textwidth}
% \centering
\vspace{-1em}
\caption{Comparison between GRExplainer and existing TGNN explanation methods.}
\label{comparison}
\resizebox{0.98\linewidth}{!}{
\begin{tabular}{lccccc}
\toprule
\multirow{2.5}{*}{\textbf{Method}} & 
\multirow{2.5}{*}{\textbf{Generality}} & 
\multirow{2.5}{*}{\textbf{Efficiency}} & 
\multicolumn{2}{c}{\textbf{User-friendliness}} \\
\cmidrule(lr){4-5}
& & & \textbf{Connectivity} & \textbf{Knowledge} \\
\midrule
\cite{xie2022explaining} & \Circle & $\gg O(E)$ & \Circle & High   \\
\cite{he2022explainer}    & \Circle & $\gg O(E)$ & \Circle & Low    \\
\cite{xia2023explaining}  & \Circle & $\gg O(E)$ & \Circle & Medium \\
\cite{chen2023tempme}     & \Circle & $\gg O(E)$ & \CIRCLE & Medium \\
\textbf{Ours}             & \CIRCLE & $O(MN)$    & \CIRCLE & Low    \\
\bottomrule
\end{tabular}
}
\scriptsize
\CIRCLE: Criterion satisfied; \Circle: Criterion not satisfied; $E$: Number of edges; $N$: Number of nodes; $M$: The largest layer size in the BFS tree; ``High'' requires knowledge of both model parameters and explanation size; ``Medium'' requires one; ``Low'' requires neither.
\vspace{-1em}
\end{wraptable}
We can find few TGNN explanation methods in the literature~\cite{xie2022explaining,he2022explainer,xia2023explaining,chen2023tempme}. Table~\ref{comparison} systematically compares them across three critical dimensions: generality, efficiency, and user-friendliness.  
Our review identifies the following limitations: 
(i) Existing explanation methods are designed specifically for either snapshot-based~\cite{xie2022explaining,he2022explainer} or event-based TGNNs~\cite{xia2023explaining,chen2023tempme}, limiting their generality across different types of dynamic graphs. 
(ii) Current methods~\cite{xia2023explaining,chen2023tempme,xie2022explaining,he2022explainer} rely on inefficient edge operations. 
% Edge-selection approaches~\cite{xia2023explaining,chen2023tempme} require processing massive edge subsets, gradient-propagation methods~\cite{xie2022explaining} must traverse the entire edge structures, and TPGM-Explainer~\cite{he2022explainer} also incurs high computational overhead due to its combinatorial perturbation and evaluation of temporal edge subsets. 
These inefficiencies would be further exacerbated in large-scale real-world networks, restricting their practicality.
(iii) Most TGNN explanation methods ~\cite{xie2022explaining,xia2023explaining,chen2023tempme} generate discrete nodes or edges rather than connected subgraphs, which makes the explanations difficult to understand. Additionally, they often require users to possess knowledge of model parameters or explanation size. Such high knowledge requirements reduce their user-friendliness for general users.

To tackle the identified issues, in this paper, we propose GRExplainer, \textit{the first universal, efficient, and user-friendly explanation method for TGNNs}. 
GRExplainer extracts node sequences as a unified feature representation and employs a Recurrent Neural Network (RNN)-based model to generate explanation subgraphs based on these sequences. 
% To the best of our knowledge, GRExplainer is the first universal, efficient, and user-friendly explanation method for TGNNs. 
Specifically, we address the following challenges:

\textit{\textbf{C1: How to explain TGNNs in a universal manner?}} 
Due to substantial differences in dynamic graph representations between snapshot-based and event-based TGNNs, developing a universal explanation method is particularly challenging. 
To address this, we abstract the input graphs of various TGNN models by extracting node sequences as a unified feature representation. Leveraging this representation, we develop a universal input method that works across diverse TGNN models, eliminating the need for model-specific input formats. \textit{Our key innovation lies in harnessing this unified representation to formulate a universal explanation method for TGNNs.}

\textit{\textbf{C2: How to make the explanation method efficient?}}
Explaining TGNNs typically involves identifying important edges, resulting in time complexity proportional to the number of edges. However, real-world networks are enriched with edges, making efficiency a major challenge.
To overcome this, we generate explanations based on structure- and time-aware node sequences, significantly reducing complexity since the number of nodes is generally much smaller than that of edges. For snapshot-based TGNNs, we adopt a Breadth-First Search (BFS)-based strategy where each newly visited node connects only to its peers at the same BFS layer and its immediate parent, minimizing node associations. For event-based TGNNs, we impose temporal constraints (i.e., nodes observed at time $t_1$ cannot connect to nodes appearing at $t_2$, where $t_1 < t_2$) to avoid redundant computation.

\textit{\textbf{C3: How to make the explanation method user-friendly?}} 
Many existing methods use non-generative techniques with hand-crafted rules~\cite{xie2022explaining,xia2023explaining,chen2023tempme}, such as setting thresholds to select important edges. This often leads to discontinuous explanations and reliance on prior knowledge. 
To address this, we propose a generative explanation model based on RNNs. RNNs can memorize previous graph states through its state propagation mechanism. Using this property, our model predicts dynamic connections between new and existing nodes, thus producing topologically continuous explanation subgraphs. The fully automated process also eliminates the need for manual input or domain expertise.

We conduct extensive experiments to evaluate GRExplainer's generality, efficiency, and user-friendliness across six datasets~\cite{yanardag2015deep,lepomaki2021retaliation,kumar2019predicting} from different fields regarding one snapshot-based TGNN (EvolveGCN~\cite{pareja2020evolvegcn}) and two event-based TGNNs (TGAT~\cite{xu2020inductive} and TGN~\cite{rossi2020temporal}). Experimental results demonstrate that GRExplainer effectively explains both types of TGNNs. For instance, it achieves a 35440\% fidelity~\footnote{Fidelity is a metric to measure the accuracy of explanations.} improvement on the Mooc dataset~\cite{kumar2019predicting}, compared with the best baseline. In terms of efficiency, GRExplainer shows a 16x speedup over the fastest baseline on the same dataset. Notably, it operates without human intervention and achieves the highest cohesiveness~\footnote{Cohesiveness is a metric to quantify the connectivity of explanations.} among the baselines, highlighting its sound user-friendliness.

% The main contributions of this paper are as follows: (i) We propose GRExplainer, the first universal explanation method capable of explaining both snapshot-based and event-based TGNNs with high efficiency and user friendliness. (ii) We improve computational efficiency by generating explanations based on node sequences and respectively adopting a BFS-based strategy and temporal constraints for two types of TGNNs, to reduce redundant computation. (iii) We enhance user-friendliness by employing a generation model to produce explanation subgraphs, greatly reducing user dependency and ensuring explanation connectivity. (iv) We perform an extensive evaluation on GRExplainer over six datasets regarding three TGNNs involving both types. Experimental results show that GRExplainer effectively explains both types of TGNNs, is much more efficient than baselines, and provides user-friendly explanations.

\section{Related Work}
\subsection{Temporal Graph Neural Networks}
Based on how dynamic graphs are represented, TGNNs can be categorized into snapshot-based~\cite{pareja2020evolvegcn,sankar2020dysat} and event-based~\cite{xu2020inductive,rossi2020temporal} models. EvolveGCN~\cite{pareja2020evolvegcn} combines Graph Convolutional Networks (GCNs)~\cite{li2021semi} with RNNs to dynamically update GCN parameters, addressing the limitations of fixed node embeddings and poor adaptability in static GNNs. DySAT~\cite{sankar2020dysat} employs two self-attention mechanisms to learn structural patterns within each snapshot and capture temporal dependencies across snapshots. TGAT~\cite{xu2020inductive} integrates time encoding with attention mechanisms to model continuous-time interactions. It captures fine-grained temporal dynamics and supports inductive learning for unseen nodes. However, TGAT may produce stale embeddings for inactive nodes. TGN~\cite{rossi2020temporal} addresses this issue via a memory-based architecture that preserves long-term dependencies and ensures up-to-date representations for inactive nodes. For a comprehensive review of TGNNs, refer to~\cite{feng2024comprehensive}. 

% \textbf{Snapshot-based TGNNs.} EvolveGCN~\cite{pareja2020evolvegcn} combines Graph Convolutional Networks (GCNs)~\cite{li2021semi} with RNNs to dynamically update GCN parameters, addressing the limitations of fixed node embeddings and poor adaptability in static GNNs. DySAT~\cite{sankar2020dysat} employs two self-attention mechanisms to learn structural patterns within each snapshot and capture temporal dependencies across snapshots. Compared with RNNs, self-attention improves efficiency through parallel processing.

% \textbf{Event-based TGNNs.} TGAT~\cite{xu2020inductive} integrates time encoding with attention mechanisms to model continuous-time interactions (i.e., events). It captures fine-grained temporal dynamics and supports inductive learning for unseen nodes. However, TGAT may produce outdated representations for nodes that are inactive for long periods. TGN~\cite{rossi2020temporal} addresses this limitation via a memory-based architecture that preserves long-term dependencies and ensures up-to-date representations for inactive nodes.

\subsection{Explanation Methods for Graph Neural Networks}
Explainability methods for static GNNs have been extensively studied and can be classified into mask-based~\cite{ying2019gnnexplainer,yuan2021subgraphX}, gradient/feature-based~\cite{baldassarre2019explainability,pope2019explainability}, decomposition-based~\cite{zhang2022gstarx,gui2023flowx}, surrogate-based~\cite{huang2022graphlime,vu2020pgm}, causal-based~\cite{wang2021causal,lin2022orphicx}, policy-based~\cite{yuan2020xgnn,funke2022zorro}, and clustering-based~\cite{shin2022prototype} methods. While effective in static settings, they may fail to produce accurate explanations for TGNNs due to the neglect of temporal information. Consequently, recent research has increasingly focused on developing explainability methods for TGNNs~\cite{xie2022explaining,he2022explainer,xia2023explaining,chen2023tempme,li2023heterogeneous,tang2023explainable}. In the following, we briefly review representative TGNN explanation methods.
 
\textbf{Snapshot-based TGNN Explanation.} DGExplainer~\cite{xie2022explaining} is a layer-wise explanation method specifically designed for GCN-Gated Recurrent Unit (GRU) models. It iteratively computes activation scores to identify important nodes and forms explanations using high-scoring ones. However, this method is computationally intensive and requires knowledge of both model parameters and explanation scale. Additionally, its node selection process inherently produces fragmented explanations.  
TPGM-Explainer~\cite{he2022explainer} combines PGM-explainer~\cite{vu2020pgm} with timeslot-based Bayesian networks to generate explanation subgraphs. While this method eliminates manual intervention by automatically modeling causal relationships using Bayesian networks, it remains computationally expensive due to edge perturbations. Moreover, the derived explanations lack structural coherence, reducing their user-friendliness.

\textbf{Event-based TGNN Explanation.} T-GNNExplainer~\cite{xia2023explaining} is a pioneering explanation method designed for event-based TGNN models. It integrates a navigator with Monte Carlo Tree Search (MCTS), where the navigator guides the sampling process of the tree search, and the sampled event groups are used to construct explanation subgraphs. However, its dependence on edge-level operations limits scalability, and the use of stopping criterion during MCTS often leads to fragmented and less user-friendly subgraphs. 
To improve coherence, TempME~\cite{chen2023tempme} introduces a graph generation approach based on motifs, incorporating three key components: domain knowledge-guided motif selection, adaptive importance scoring, and selective motif aggregation. Nevertheless, TempME heavily relies on domain expertise for motif identification, reducing its accessibility to users. Moreover, its edge-based operations compromises efficiency.

\section{Preliminaries and Problem Formulation}
\subsection{Preliminaries}
\noindent \textbf{Definition 1: Snapshot Graph.}
It represents a dynamic graph as a sequence of static snapshots $\{G_t | n = 1, 2, \dots, T\}$, where $T$ is the number of snapshots. Each snapshot ${G}_t = (V_t, E_t)$ captures the graph structure observed at the $t$-th timeslot, where ${V}_t$ and ${E}_t$ denotes the sets of nodes and edges, respectively. The parameter $T$ relates to the frequency of graph observation.

\noindent \textbf{Definition 2: Event Graph.}
It represents a dynamic graph as a stream of events $\{e_i | i=1,2,\cdots\}$, where each event incrementally updates the graph. An event may involve an edge addition or deletion, or a node attribute update, and is timestamped with $t_i$ to preserve fine-grained temporal information.

\subsection{Problem Formulation}
Let $f$ be a well-trained target model used to predict whether an interaction edge $e$ occurs between nodes $v_i$ and $v_j$ at timestamp $t$, based on time-aware node representations. The model outputs either probabilities or logits. The objective of an explanation method is to identify a subset of historical interactions that significantly influence the target model $f$'s prediction of future interactions. This subset of important interactions is known as an explanation. Formally, let $Y_f[e]$ denote the prediction of interaction $e$ made by the target model $f$. The explanation task is formulated as maximizing the mutual information between the explanation and the original model prediction:
\begin{equation}
\label{problem_formula}
\underset{|G_{\text{sub}}| \leq K_{\text{sub}}}{\arg\max} \, I(Y_f[e]; G_{\text{sub}}) \Leftrightarrow \underset{|G_\text{sub}| \leq K_{\text{sub}}}{\arg\min} \, -\sum_{c \in \mathcal{C}} \mathbbm{1}(Y_f[e] = c) \log(f(G_{\text{sub}})[e]),
\end{equation}
where $I(\cdot,\cdot)$ denotes the mutual information function, $e$ is the interaction to be explained, and $G_{\text{sub}}$ denotes the explanation constructed by several important historical interactions before timestamp $t$. $f(G_{\text{sub}})[e]$ represents the prediction probability of $e$ given the explanation $G_{\text{sub}}$, and $K_{\text{sub}}$ limits the explanation size. $\mathcal{C}$ is the set of prediction classes, e.g., $\mathcal{C} = \{0, 1\}$ in binary classification. We refer to Appendix~\ref{appendix_problem} for the theoretical proof of Eq.~\ref{problem_formula}.

\section{GRExplainer Design}
\subsection{Overview}
\begin{figure*}[tbp]
	\centering
	\includegraphics[width=1.0\textwidth]{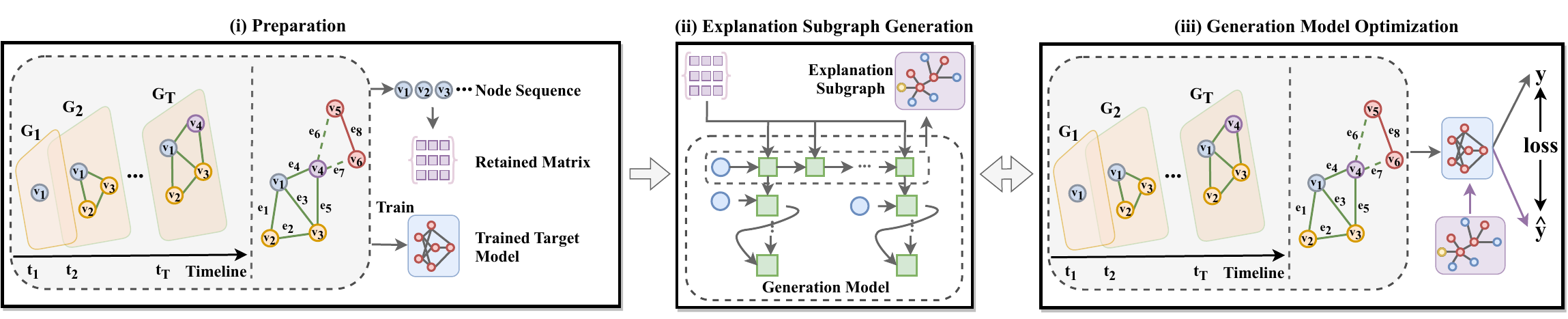}
        \vspace{-4mm}
	\caption{GRExplainer overview.}
	\label{all_graph}
        \vspace{-3mm}
\end{figure*}

As illustrated in Fig.~\ref{all_graph}, GRExplainer consists of three key stages: preparation, explanation subgraph generation, and generation model optimization. In the preparation stage, we process both snapshot and event graphs and train target TGNN models for tasks like link prediction. To ensure generality across different types of TGNNs, we abstract the input graphs into node sequences and compute their retained matrices. For efficiency, BFS is used for snapshot graphs, while temporal ordering is considered for event graphs. In the explanation subgraph generation stage, an RNN-based generation model takes the retained matrices as input to automatically generate explanation subgraphs. Finally, during the optimization stage, the generation model iteratively refines the explanation subgraphs by minimizing prediction discrepancies between the target model's outputs on the generated subgraphs and the original graphs.
% This approach provides two key advantages: (1) ensuring explanation coherence through sequential pattern learning, and (2) eliminating manual intervention via self-contained generation.
% In the generation model optimization phase, we input both generated explanation subgraphs and original datasets in the preparation phase into the target model, then optimize the generation model by minimizing prediction discrepancies. This iterative process continues until convergence, producing the final explanation subgraph.

\subsection{Preparation} 
In this stage, we train TGNN models on various datasets and prepare input data for explanation generation. To unify input formats, we apply preprocessing strategies tailored to each type of dynamic graph. Focusing on link prediction, we extract edge-centered subgraphs to reduce noise and improve efficiency, as only a subset of the graph is relevant for each prediction.

\textbf{Snapshot Graph.} Given a snapshot \( G_t = (V_t, E_t) \) and a target edge \( e = (v_i, v_j) \in E_t \) to be explained, we construct a subgraph centered around the \( k \)-hop neighborhoods of both endpoints. Let \( N_k(v) \) denote the set of nodes reachable from node \( v \) within \( k \) hops. The subgraph is induced by the node set $V_{\text{subgraph}} = \{ v_i, v_j \} \cup N_k(v_i) \cup N_k(v_j)$ and the adjacency matrix \( A_{\text{subgraph}} \). To unify input formats with efficiency in mind, we extract a node sequence via Breadth-First Search (BFS) starting from a randomly selected node \( v_1 \in V_{\text{subgraph}} \), yielding  $V_{\text{sequence}} = [v_1, \dots, v_n]$. Compared to arbitrary node permutations, BFS-based sequences significantly reduce the search space, since BFS traversal inherently limits connections to predecessors.

Specifically, for any BFS sequence \( [v_1, \dots, v_n] \) in \( G_t = (V_t, E_t) \), if \( (v_i, v_{j-1}) \in E_t \) but \( (v_i, v_j) \notin E_t \) where \( i < j \leq n \), then \( (v_{i'}, v_{j'}) \notin E_t \) for all \( 1 \leq i' \leq i \) and \( j \leq j' < n \) (see the proof in Appendix~\ref{appendix_bfs}). This property implies that each node can only connect to its parent or to nodes at the same BFS layer, enabling efficient connectivity constraints. 

Based on this insight, we define a retained matrix \( A_{\text{retained}} \) that filters connections based on a fixed number \( M \) of preceding nodes in the BFS sequence:
\begin{equation}
A_{\text{retained}} =
\begin{cases}
1, & \text{if } (v_r, v_s) \in A_{\text{subgraph}} \text{ and } \text{rank}(v_r) \leq \min(M, \text{rank}(v_s)) \\
0, & \text{otherwise}
\label{A_{retained}}
\end{cases}
.
\end{equation}
Here, \( \text{rank}(v) \) denotes the position of node \( v \) in the sequence, and \( M \) is defined as:
\begin{equation}
M = O \left( \max_{d=1}^{\text{diam}(G_t)} \left| \left\{ v_i \mid \text{dist}(v_i, v_1) = d \right\} \right| \right),
\end{equation}
where \( \text{dist}(v_i, v_1) \) is the shortest-path distance between \( v_i \) and the starting node \( v_1 \), and \( \text{diam}(G_t) \) denotes the diameter of the graph. This bound on \( M \) ensures that structural dependencies can be captured with limited context, facilitating subsequent steps (e.g., training of the generative model).

\textbf{Event Graph.} 
For event graphs, we first identify a sub-event group ${e}_{\text{subgraph}}$ around the target edge. Nodes are then sorted by timestamps to form the sequence $[v_1, \dots, v_n] = \text{sort}(\{v_i \mid t_i\})$. Similar to the snapshot case, we retain only the connections to the previous $M$ nodes with $\text{rank}(v)$ being the temporal position of $v$, yielding the retained matrix $A_{\text{retained}}$.

\textbf{University Discussion.} 
GRExplainer extracts node sequences as a unified feature representation for both snapshot graphs and event graphs. This representation captures structural or temporal dependencies within the graph and serves as a model-agnostic input for efficient explanation generation.

\subsection{Explanation Subgraph Generation}
In this stage, we generate explanation subgraphs using a novel generation model inspired by GraphRNN~\cite{you2018graphrnn}. Unlike GraphRNN's general-purpose two-component RNN architecture, our model implements a specialized one-to-many RNN structure for TGNN explanation. This design enables sequential prediction of node connections by learning from established connectivity patterns, thereby enhancing structural dependency modeling -- a critical requirement for explanation generation.

\begin{wrapfigure}{r}{0.48\textwidth}
    \centering
    \vspace{-2mm}
    \includegraphics[width=\linewidth]{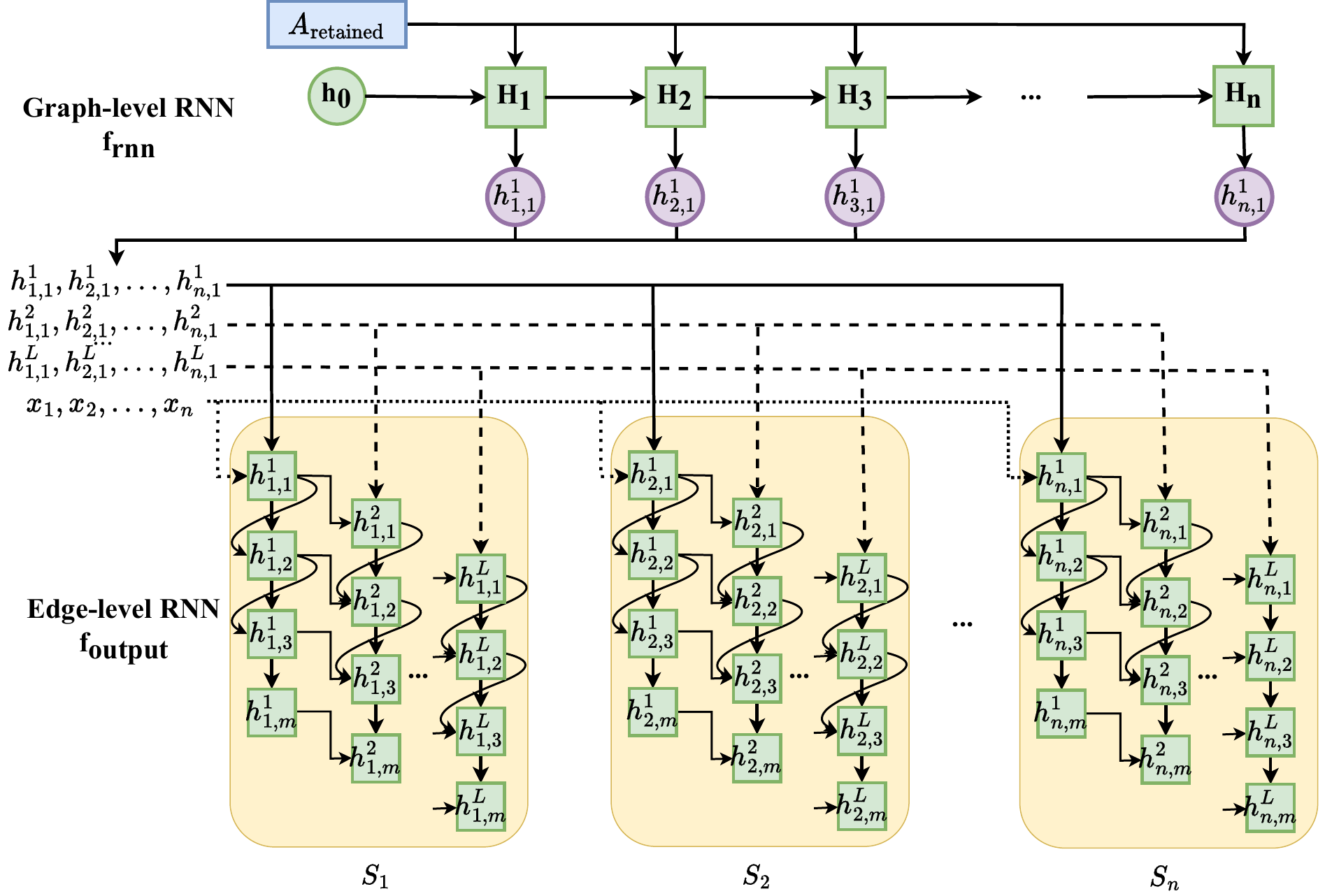} % 自动适应栏宽
    \vspace{-4mm} % 微调间距
    \caption{Generation model architecture.} % 优化标题表述
    \label{fig:generation_model}
    \vspace{-4mm}
\end{wrapfigure}
As shown in Fig.~\ref{fig:generation_model}, our model consists of a graph-level RNN (${{f}_\text{rnn}}$) that generates nodes while maintaining graph states, and an edge-level RNN (${{f}_\text{output}}$) that predicts connections for newly generated nodes. We adopt RNNs, specifically GRUs~\cite{chung2014empirical}, due to their ability to model sequential dependencies through hidden states, which effectively capture historical structural context. Furthermore, the autoregressive nature of RNNs aligns well with our objective of generating coherent explanation subgraphs step by step.

In the generation model, each node $v_i$ is associated with an adjacency vector $S_{i}$, which specifies its connections to all preceding nodes in the sequence. This vector is modeled as a product of independent Bernoulli trials, each conditioned on the previous graph state, capturing the probability of an edge between $v_i$ and any preceding node $v_j$ $(j<i)$. Formally, this is defined as: 
\begin{equation} p(S_{i} \mid S_{<i}) = \prod_{j=1}^{i-1} p(S_{i,j} \mid S_{i,<j}, S_{<i}),
\label{P} 
\end{equation} 
where $S_{i,j}$ is a binary scalar indicating whether there is an edge between node $v_i$ and node $v_j$.

At the beginning of generation, we initialize a binary vector (concatenated with \( A_{\text{retained}} \)) and \( h_0 \) as input to the graph-level RNN, since the first node has no preceding nodes. The resulting outputs $\{h_{i,1}^1\}_{i=1}^n$ represent the evolving graph states and serve as initial hidden states for the edge-level RNN. The edge-level RNN also receives randomly initialized binary inputs $\{x_i\}_{i=1}^n$, along with hidden states $\{h_{i,1}^l\}_{i=1}^n$ for $l = 2, \dots, L$, if multiple layers are used. It then outputs the adjacency vector $S_{i}$ for each node $v_i$, which is passed through a Multilayer Perceptron (MLP) to produce the edge existence probability vector $P_{\text{edge}}^{(i)} \in \mathbb{R}^{i-1}$, where each entry denotes the probability of an edge between $v_i$ and $v_j$ $(j<i)$. The overall process is as follows:
\begin{equation}
    \{h_{i,1}^1\}_{i=1}^n = f_{\text{rnn}}(A_{\text{retained}}, h_0),
    \label{rnn}
\end{equation}
\begin{equation}
    S_i = f_{\text{output}}(\{h_{i,j}^l \mid j=1,\dots,m; \ l=1,\dots,L\}, x_i),
    \quad \forall i \in \{1, \dots, n\},
    \label{output}
\end{equation}
\begin{equation}
    P_{\text{edge}}^{(i)} = \text{MLP}(S_i).
    \label{edge_p}
\end{equation}

Based on \( P_{\text{edge}}\), we generate the explanation subgraph ${G}_\text{sub}$ by retaining edges in $A_{\text{subgraph}}$.
% The procedures of generating explanations for snapshot-based and event-based TGNNs are shown in Algorithm~\ref{algorithm_snapshot} and Algorithm~\ref{algorithm_event}, respectively. 

\textbf{University Discussion.} 
For snapshot-based models, GRExplainer aggregates the generated subgraphs across snapshots to form the final explanation ${G_{\text{sub}}}$, while for event-based models, the generated ${G_{\text{sub}}}$ directly constitutes the explanation. The pseudo-code of GRExplainer is given in Appendix~\ref{algorithm}.

\subsection{Generation Model Optimization} 
In this stage, we optimize the generation model to produce the final explanation subgraph. A modified binary cross-entropy loss is used to minimize the discrepancy between predictions on the explanation subgraph and the original dynamic graph. The loss function is defined as:
\begin{equation}
L = \lambda_{\text{size}} \sum A_{\text{sub}} - \lambda_{\text{weight}} |\hat{y} - y|.
\label{loss}
\end{equation}

The first term, weighted by $\lambda_{\text{size}}$, penalizes the size of the subgraph via its adjacency matrix $A_{\text{sub}}$, promoting concise and localized explanations. The second term, scaled by $\lambda_{\text{weight}}$, preserves fidelity by aligning the explanation’s prediction $\hat{y}$ with that of the original graph $y$.

\textbf{Time Complexity.}
GRExplainer's time complexity is dominated by the edge-level RNN, with optimization reducing it to $O(MN)$, where $N$ is the number of nodes in the graph, and $M$ is the maximum number of nodes in any single layer during BFS traversal. This localized exploration avoids full-graph traversal while preserving structural integrity.
% Through this optimization strategy, we significantly lower the time complexity, offering a notable advantage, especially when handling large-scale graphs.

\section{Experiments}
We aim to answer the following research questions to evaluate the effectiveness of GRExplainer. 
\textbf{RQ1:} Can GRExplainer provide explanations for both snapshot-based and event-based TGNNs, and how does its accuracy compare to state-of-the-art baselines?
\textbf{RQ2:} How do different components of GRExplainer contribute to its performance?
\textbf{RQ3:} Is the explanation efficiency of GRExplainer superior to that of baselines?
\textbf{RQ4:} How user-friendly is GRExplainer compared to baselines?
We also analyze the effect of hyperparameters in Appendix~\ref{appendix_hyparameters}.

\subsection{Experimental Settings} \label{experimental_settings}
\textbf{Datasets.} 
We adopt six real-world datasets spanning diverse domains, including social networks, financial networks, and online education: Reddit-Binary~\cite{yanardag2015deep}, Bitcoin-Alpha~\cite{lepomaki2021retaliation}, Bitcoin-OTC~\cite{lepomaki2021retaliation}, Reddit~\cite{kumar2019predicting}, Wikipedia~\cite{kumar2019predicting}, and Mooc~\cite{kumar2019predicting}. These datasets encompass both snapshot and event graphs, enabling a comprehensive evaluation across different temporal settings. Detailed dataset statistics, descriptions, and preprocessing procedures are provided in Appendix~\ref{appendix_dataset}.

\textbf{Target Models.}
We select three widely used TGNNs as target models for explanation. Specifically, EvolveGCN~\cite{pareja2020evolvegcn} is chosen to represent snapshot-based TGNNs, while TGAT~\cite{xu2020inductive} and TGN~\cite{rossi2020temporal} are selected for event-based TGNNs. Despite this, GRExplainer is universal and can be applied to any TGNN model.

\textbf{Baseline Methods.} 
We compare GRExplainer with representative explanation methods for both snapshot-based and event-based TGNNs. For snapshot-based TGNNs, due to the lack of implementation details of DGExplainer~\cite{xie2022explaining} and TPGM-Explainer~\cite{he2022explainer}, we adapt GNNExplainer~\cite{ying2019gnnexplainer} as a baseline. GNNExplainer is a strong baseline known to outperform several non-learning-based methods~\cite{pope2019explainability}. We apply it to each snapshot independently, generate explanations for individual snapshots, and aggregate them to form the final explanation. For event-based TGNNs, we select two state-of-the-art explanation methods, T-GNNExplainer~\cite{xia2023explaining} and TempME~\cite{chen2023tempme}, as baselines. 

\textbf{Evaluation Metrics.}
To evaluate the accuracy of explanation methods, we adopt two metrics: \textit{Fidelity+ (FID+)} and \textit{the Area Under the Fidelity-Sparsity Curve (AUFSC)}. FID+ quantifies the impact of important features—identified by an explanation method—by measuring the change in prediction probability when these features are masked~\cite{pope2019explainability}. AUFSC summarizes FID+ scores at different sparsity levels. Herein, sparsity refers to the proportion of retained edges or nodes (selected based on their importance scores) in the explanation subgraph. Higher FID+ and AUFSC indicate better explanation accuracy. To enable intuitive comparison, we report the \textit{Best FID+}, which indicates the highest FID+ score achieved by an explanation method without sparsity limitations.

To assess the efficiency of explanation methods, we measure the \textit{Runtime} taken to generate explanations. A lower value indicates higher efficiency.

To evaluate the connectivity of explanations, we adopt the \textit{Cohesiveness} metric~\cite{chen2023tempme}, which quantifies the spatial and temporal coherence of interactions within the explanation subgraph. A higher cohesiveness score indicates better connectivity in the explanation.

\textbf{Implementation Details.}
We implemented GRExplainer and baselines using Pytorch on a server equipped with an Intel Xeon Silver 4216 CPU and an RTX 3090 GPU. For TGAT and TGN, we adopted their original two-layer attention architectures and data splitting strategies. EvolveGCN was implemented with a two-layer GCN. Considering memory constraints, we used 3, 4, and 6 snapshots for Reddit, Bitcoin-Alpha, and Bitcoin-OTC, respectively. During explanation generation, the maximum number of training epochs was set to 50, with early stopping triggered if AUFSC does not improve for 10 consecutive epochs. To reduce randomness, we generated multiple explanation instances and reported their average performance. Other details are given in Appendix~\ref{appendix_detail}.

\subsection{Performance Comparison (RQ1)}
Tables~\ref{table:snapshot} and~\ref{table:event} present the quantitative results. GRExplainer outperforms GNNExplainer in explaining snapshot-based TGNNs. On the three snapshot graph datasets, it achieves AUFSC gains of 60.3\%, 7.4\%, and 14.3\%, and FID+ gains of 194.4\%, -0.6\%, and 1.7\% for the EvolveGCN model. The modest gains on Bitcoin-Alpha and Bitcoin-OTC are likely due to their small size and simple topology, which restrict structural expressiveness and hinder informative subgraph generation.
On the event graph datasets, GRExplainer significantly surpasses T-GNNExplainer and TempME in explaining TGAT, with AUFSC gains up to 283.1\% and FID+ gains reaching 10125\%, compared with the best baseline. Its performance remains strong for TGN, achieving AUFSC gains up to 299.8\% and FID+ gains of 35440\%. These results highlight GRExplainer’s strength in capturing complex decision patterns in temporally rich data. However, on the Reddit dataset, GRExplainer underperforms TempME in AUFSC. This may stem from the incompatibility between TempME's motif-based subgraphs and the TGN model, which results in low prediction accuracy. In this case, the prediction score of a sparse graph tends to be close to that of the original graph, yielding higher AUFSC than our method.

\begin{table}[tb]
    \centering
    \caption{Performance comparison of GRExplainer and a baseline for explaining snapshot-based TGNNs. The best result is in \textbf{bold}.}
    \label{table:snapshot}
    \vspace{0.5em}
    \resizebox{0.85\linewidth}{!}{
    \begin{tabular}{llcccccc}
        \toprule
        \multirow{2.5}{*}{Target Model} & \multirow{2.5}{*}{Method} 
        & \multicolumn{2}{c}{Reddit-Binary} 
        & \multicolumn{2}{c}{Bitcoin-Alpha} 
        & \multicolumn{2}{c}{Bitcoin-OTC} \\
        \cmidrule(lr){3-4} \cmidrule(lr){5-6} \cmidrule(lr){7-8}
        & & Best FID+ & AUFSC & Best FID+ & AUFSC & Best FID+ & AUFSC \\
        \midrule
        \multirow{2}{*}{EvolveGCN} 
        & GNNExplainer & -0.072 & -0.063 & \textbf{0.624} & 0.148 & 0.875 & 0.488 \\
        & \textbf{GRExplainer} & \textbf{0.068} & \textbf{-0.025} & {0.620} & \textbf{0.159} & \textbf{0.890} & \textbf{0.558} \\
        \bottomrule
    \end{tabular}}
    \vspace{-3mm}
\end{table}

\begin{table}[tb]
    \centering
    \caption{Performance comparison of GRExplainer and baselines for explaining event-based TGNNs. The best result is in \textbf{bold} and second best is \underline{underlined}.}
    \label{table:event}
    \vspace{0.5em}
    \resizebox{0.85\linewidth}{!}{
    \begin{tabular}{llcccccc}
        \toprule
        \multirow{2}{*}{Target Model} & \multirow{2}{*}{Method}
        & \multicolumn{2}{c}{Reddit} & \multicolumn{2}{c}{Wikipedia} & \multicolumn{2}{c}{Mooc} \\
        \cmidrule(lr){3-4} \cmidrule(lr){5-6} \cmidrule(lr){7-8}
        & & Best FID+ & AUFSC & Best FID+ & AUFSC & Best FID+ & AUFSC \\
        \midrule
        \multirow{3}{*}{TGAT}
        & T-GNNExplainer & -0.066 & -1.738 & -0.034 & -0.992 & -0.029 & -0.590 \\
        & TempME         &  \underline{0.235} & \underline{-0.650} &  \underline{0.001} & \underline{-0.718} &  \underline{0.008} & \underline{-0.178} \\
        & \textbf{GRExplainer} & \textbf{0.753} & \textbf{-0.412} & \textbf{0.567} & \textbf{-0.252} & \textbf{0.818} & \textbf{0.326} \\
        \midrule
        \multirow{3}{*}{TGN}
        & T-GNNExplainer & -0.142 & -2.653 & -0.122 & -0.696 & \underline{0.005} & -1.123 \\
        & TempME         &  \underline{0.029} & \textbf{-0.714} & \underline{-0.003} & \underline{-0.429} & 0.004 & \underline{-0.479} \\
        & \textbf{GRExplainer} & \textbf{0.073} & \underline{-2.308} & \textbf{0.226} & \textbf{-0.250} & \textbf{1.777} & \textbf{0.957} \\
        \bottomrule
    \end{tabular}}
    \vspace{-3mm}
\end{table}

\begin{figure}[htbp]
    \centering
    \begin{minipage}[t]{0.48\textwidth}
        \centering
        \includegraphics[width=\linewidth]{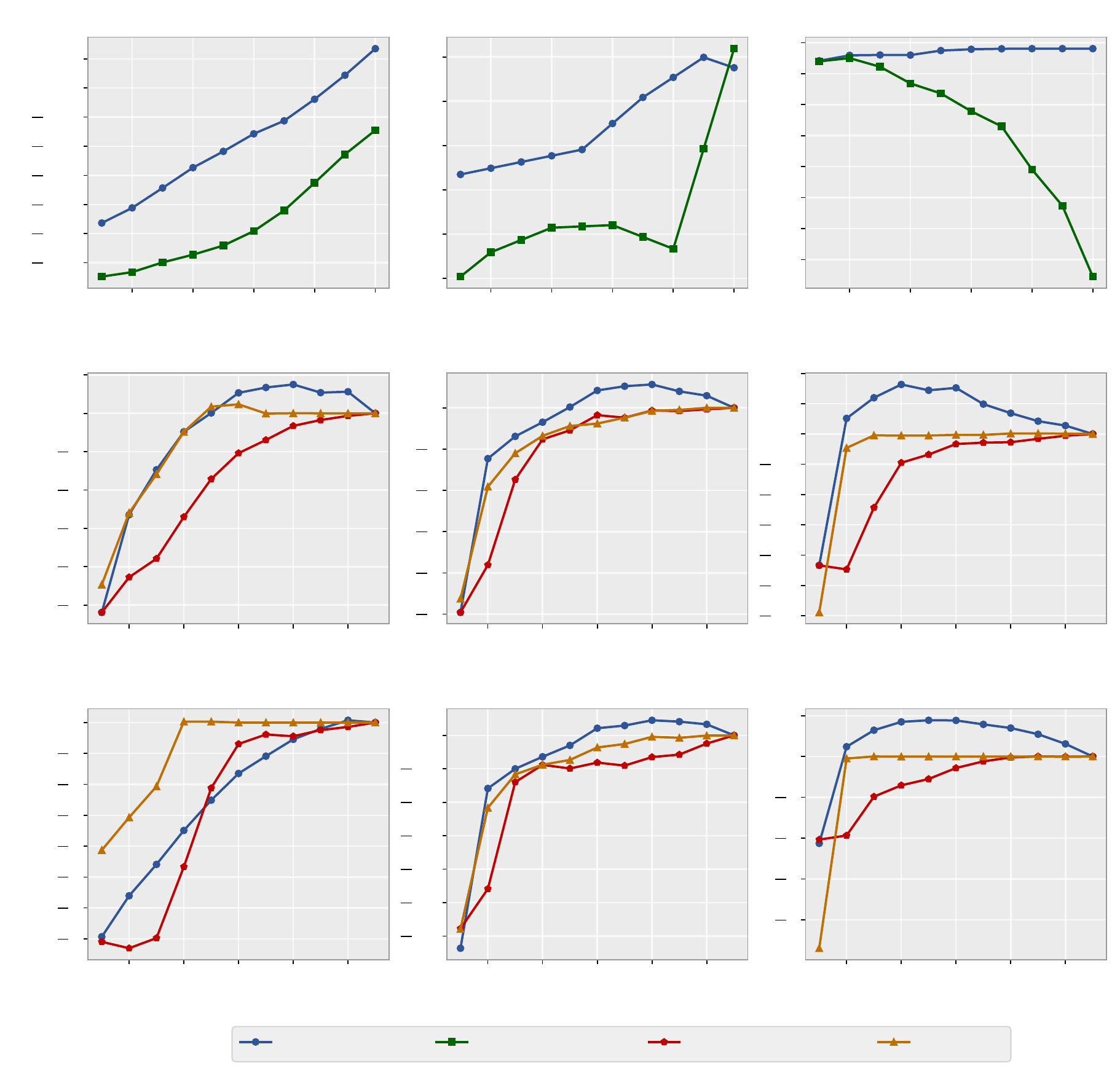}
        \caption{Fidelity-Sparsity comparison of GRExplainer and baselines.}
        \label{aufsc}
    \end{minipage}
    \hfill
    \begin{minipage}[t]{0.48\textwidth}
        \centering
        \includegraphics[width=\linewidth]{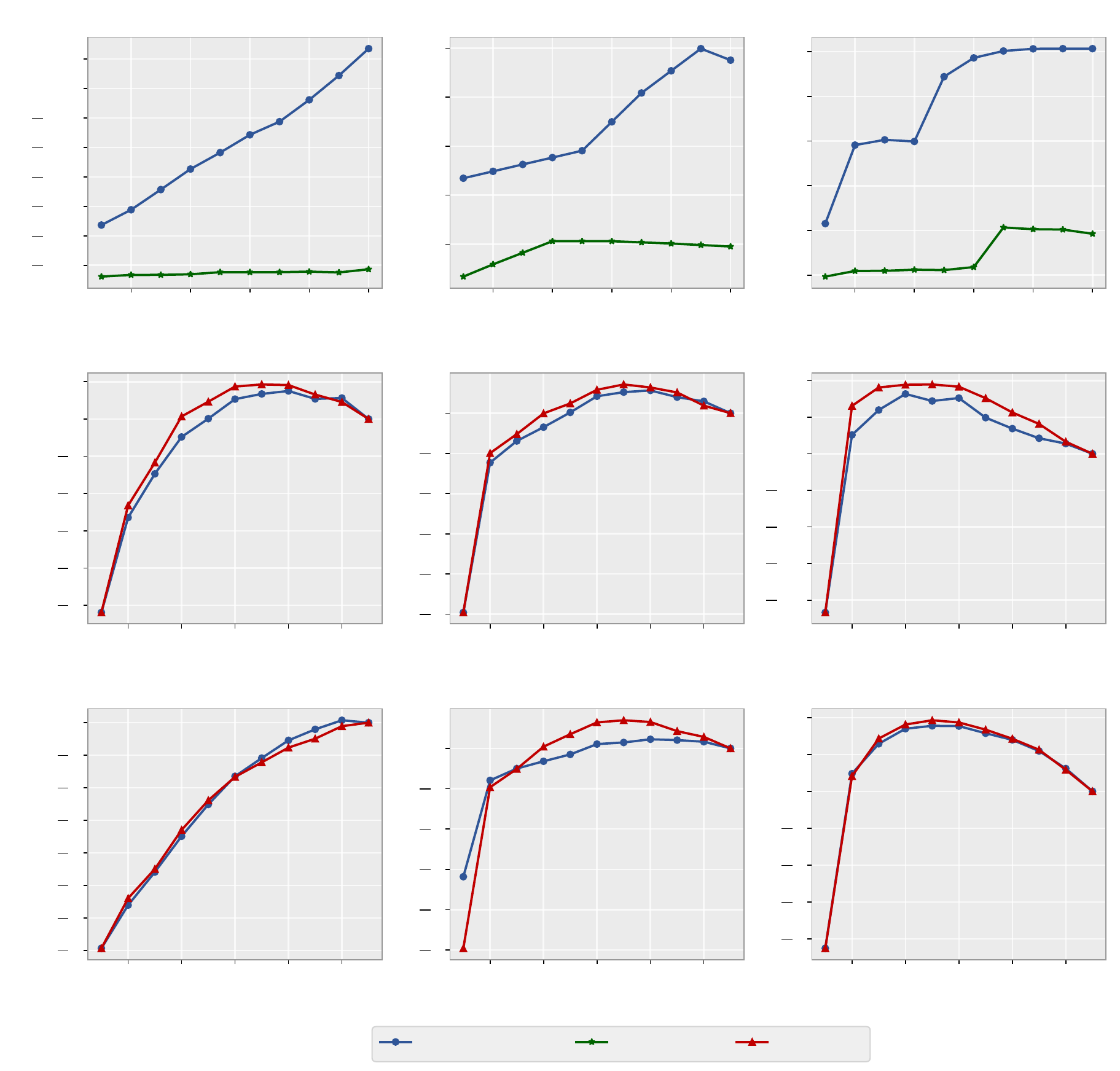}
        \caption{Fidelity-Sparsity comparison with and without temporal information.}
        \label{ablation}
    \end{minipage}
\end{figure}

Fig.~\ref{aufsc} further compares GRExplainer with baselines under various sparsity levels. From the figure, we can observe that GRExplainer consistently achieves the highest Best FID+ and AUFSC in most cases, confirming its effectiveness and generality across diverse TGNN models.
% For explaining snapshot-based TGNNs, the results show that GRExplainer has superiority over the baseline. In the case of event-based TGNNs, GRExplainer outperforms T-GNNExplainer across all sparsity levels for TGAT and also surpasses TempME on the Wikipedia and Mooc datasets. On Reddit, GRExplainer performs comparably to TempME at low sparsity levels (0.0–0.4) but outperforms it at higher sparsity levels (0.4–1.0). Additionally, GRExplainer significantly outperforms both baselines for TGN on Wikipedia and Mooc, but on Reddit, while it exceeds T-GNNExplainer in AUFSC, it does not surpass TempME, likely due to TGN's lower prediction accuracy on this dataset.

% Overall, GRExplainer consistently achieves superior performance in terms of the FID+ metric across all datasets, and outperforms baselines in AUFSC on most datasets. These results confirm the effectiveness and generality of GRExplainer in explaining diverse TGNN models under varying sparsity conditions.

\subsection{Ablation Study (RQ2)}
To validate the necessity and rationality of each component, we construct two variants: 
(i) ``w/o BFS'' (for snapshot graphs): This variant omits the BFS-based ordering and employs a randomly ordered node sequence to construct the generation model's input. 
(ii) ``w/o Time'' (for event graphs): This variant ignores temporal information and instead uses a randomly ordered node sequence as input to the generation model. 
An analysis of the sparsity regularization term is provided in Appendix~\ref{appendix_ablation}.

Table~\ref{table:ablation} validates the necessity of the BFS method. Using BFS, GRExplainer improves explanation efficiency by 7.1\%, 22.3\%, and 7.9\% on the Reddit-Binary, Bitcoin-Alpha, and Bitcoin-OTC datasets, respectively, significantly reducing explanation time. The first row of Fig.~\ref{ablation}
further shows that BFS enhances explanation accuracy by optimizing node sequences. Overall, the BFS method improves both efficiency and accuracy, making it a practical and scalable approach for explaining predictions on large-scale snapshot graphs.

Table~\ref{table:ablation} also shows that incorporating temporal information boosts explanation efficiency, e.g., by 48.0\% and 71.7\% on the Mooc dataset for TGAT and TGN, respectively. However, the last two rows of Fig.~\ref{ablation} indicate that its impact on explanation accuracy is limited. While removing temporal order may slightly improve accuracy by increasing node combination diversity, it often results in incoherent explanations due to misalignment between these combinations and the actual event sequence.
% Removing temporal order may slightly improve accuracy by increasing the diversity of node combination. However, these combinations often lead to incoherent explanations due to misalignment with actual event sequences.

\begin{table}[tbp]
    \centering
    \footnotesize
    \caption{Runtime (seconds) comparison of GRExplainer and its variants.}
    \label{table:ablation}
    \vspace{0.5em}
    \resizebox{\linewidth}{!}{
    \begin{tabular}{lccccccccc}
        \toprule
            \multirow{2.5}{*}{Method} & \multicolumn{3}{c}{EvolveGCN} & \multicolumn{3}{c}{TGAT} & \multicolumn{3}{c}{TGN} \\
        \cmidrule(lr){2-4} \cmidrule(lr){5-7} \cmidrule(lr){8-10}
        & Reddit-Binary & Bitcoin-Alpha & Bitcoin-OTC & Reddit & Wikipedia & Mooc & Reddit & Wikipedia & Mooc \\
        \midrule
        \textbf{GRExplainer} & \textbf{197.93} & \textbf{7.75} & \textbf{8.32} & \textbf{3716.90} & \textbf{2488.00} & \textbf{1752.76} & \textbf{3353.15} & \textbf{2192.60} & \textbf{885.42} \\
        - w/o BFS & 213.11 & 9.98 & 9.03 & -- & -- & -- & -- & -- & -- \\
        - w/o Time & -- & -- & -- & 3734.73 & 2509.64 & 3368.19 & 3380.05 & 2563.28 & 3129.16 \\
        \bottomrule
    \end{tabular}
    }
    \vspace{-3mm}
\end{table}

\subsection{Efficiency Analysis (RQ3)}
\begin{wrapfigure}{r}{0.6\textwidth}
        \vspace{-0.5cm}
        \centering
        \includegraphics[width=\linewidth]{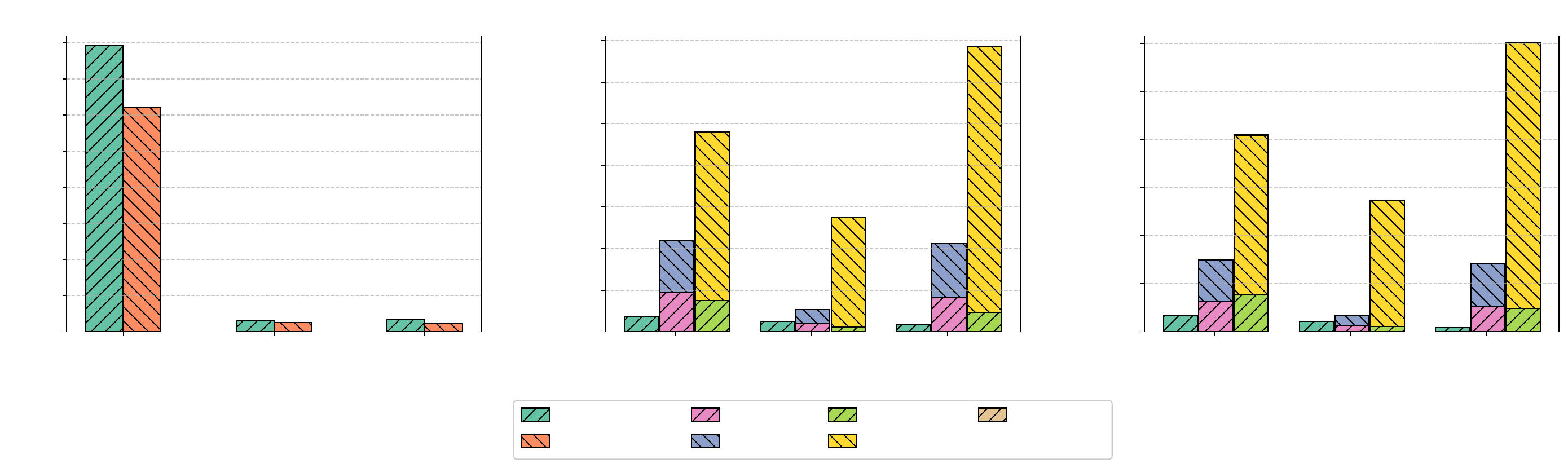}
        \vspace{-0.4cm}
        \caption{Efficiency comparison of GRExplainer and baselines.}
       \label{effictive}
       \vspace{-0.3cm}
\end{wrapfigure}
Fig.~\ref{effictive} presents the runtime comparison of GRExplainer and baselines for explaining snapshot-based and event-based TGNN models, respectively. For EvolveGCN, GRExplainer achieves comparable efficiency to GNNExplainer while offering higher accuracy. For event-based TGNN models, T-GNNExplainer's runtime includes TGNNExplainer-PT (pretrained navigator for edge scoring) and TGNNExplainer-E (explanation generation). TempME's runtime involves TempME-Motif (motif extraction), TempME-T (generator training), and TempME-E (explanation generation).
As shown in Fig.~\ref{effictive}, GRExplainer exhibits a clear efficiency advantage over baselines. For TGAT, it reduces runtime by up to 91.7\% over T-GNNExplainer and 97.4\% over TempME across three datasets. Similar improvements are observed for TGN, with reductions of up to 93.8\% and 98.5\%, respectively.
We attribute GRExplainer's high efficiency to its generative design, which ensures that the computational complexity scales with the number of nodes rather than edges. Two strategies for generating structure- and time-aware node sequences further eliminate redundant computation, enhancing its scalability for explaining TGNN models in large-scale networks.

While TempME incurs high initial costs for motif extraction and training, it benefits from this process to explain instances with similar characteristics (i.e., class-level explanations~\cite{li2025can}). In contrast, GRExplainer produces instance-level explanations, achieving high efficiency for each instance.
% Further discussion is given in Appendix~\ref{limitation}.
% TempME requires a longer time for motif extraction and training, once the training is complete, it benefits from this pre-processing in subsequent explanations. In contrast, GRExplainer explains one instance at a time, showing higher efficiency for individual explanations. 
% Therefore, GRExplainer demonstrates a clear advantage in the efficient explanation of individual instances, while TempME is better suited for scenarios where training results are reused across multiple instances.

% \begin{figure}[tbp]
%     \centering
%     %\vspace{-3mm}
%     \subfigcapskip=-3pt

%     % \setcounter{subfigure}{0}
%     %\hspace{-2mm}
%     \subfigure{\includegraphics[width=0.25\textwidth]{figure/model_comparison_tgn_baseline_effitive_2_notime_new_latest_1.pdf}\label{rq3_emb1}}
%     \vspace{-4mm}
%     \caption{Comparison of explanation efficiency between GRExplainer and baselines on EvolveGCN model.}
%     \label{EvolveGCN_Time}
%     %\vspace{-3mm}
% \end{figure}

% \begin{figure}[tbp]
%     \centering
%     %\vspace{-3mm}
%     \subfigcapskip=-3pt

%     % \setcounter{subfigure}{0}
%     % \hspace{-2mm}
%     \subfigure{\includegraphics[width=0.23\textwidth]{figure/comparison_plot_tgat_new.pdf}\label{rq3_emb1}}
%     \subfigure{\includegraphics[width=0.23\textwidth]{figure/comparison_plot_tgn_new.pdf}\label{rq3_emb2}}
%     \vspace{-4mm}
%     \caption{Comparison of explanation efficiency between GRExplainer and baselines on TGAT and TGN models.}
%     \label{TGAT_TGN_Time}
%     %\vspace{-3mm}
% \end{figure}

\subsection{User-Friendliness Analysis (RQ3)}
As previously discussed, GRExplainer requires minimal prior knowledge (an important aspect of user-friendliness), unlike baselines that rely on model parameters or predefined explanation sizes. In this subsection, we further examine user-friendliness by evaluating explanation connectivity.

\begin{wraptable}{r}{0.45\textwidth}
    \centering 
    \vspace{-0.3cm}
    \caption{Comparison of explanation cohesiveness.}
    \label{table:cohesiveness}
    %\vspace{0.5em}
    \resizebox{\linewidth}{!}{
    \begin{tabular}{llc}
        \toprule
        Setting & Method & Cohesiveness \\
        \midrule
        \multirow{2}{*}{\makecell[l]{EvolveGCN-\\Bitcoin-Alpha}} 
            & \textbf{GRExplainer} & \textbf{-2.833} \\
            & GNNExplainer         & -2.845 \\
        \midrule
        \multirow{3}{*}{TGAT-Mooc} 
            & \textbf{GRExplainer} & \textbf{-4.644} \\
            & T-GNNExplainer       & -4.655 \\
            & TempME               & -5.081 \\
        \bottomrule
    \vspace{-0.5cm}
    \end{tabular}
    }
\end{wraptable}
% To quantify this property, we adopted the cohesiveness metric, which reflects the spatiotemporal consistency of the explanation results. For a snapshot that lacks temporal variation, the ${\cos\left( \frac{|t_i - t_j|}{\Delta T} \right)}$ in Eq.~\ref{eq:log-cohensive} is set to 1, and the average of multiple explanations is taken as the final cohesiveness value. 
% For evaluation, we selected two representative TGNN models: the snapshot-based EvolveGCN and the event-based TGAT, using the Bitcoin-Alpha and Mooc datasets, respectively. These two datasets were selected for their relatively modest scale, which enables more tractable visualization and qualitative examination of the generated explanations. 
Table~\ref{table:cohesiveness} reports the cohesiveness scores of various explanation methods across different TGNN models and datasets. The results show that GRExplainer achieves the highest scores in both settings, indicating that its generated explanations are more temporally and spatially coherent than those of the baselines. This coherence facilitates users' understanding of the model's behavior. Detailed visualizations of the derived explanations are provided in Appendix~\ref{appendix_visualization}.
% In contrast, baseline methods yield lower cohesiveness scores with greater fluctuations in time, reflecting insufficient consistency and limited user-friendliness.

% To further illustrate this, Fig.~\ref{explainer_sub_tgat} visualizes explanation subgraphs for TGAT predictions on the Mooc dataset, generated by GRExplainer, T-GNNExplainer, and TempME. Although all methods produce relatively coherent explanations, GRExplainer's explanations are more localized around the edges to be explained, thus being more human-intelligible. In summary, GRExplainer offers superior user-friendliness by generating cohesive explanations without prior knowledge, making it an accessible choice for real-world applications.

\section{Conclusion}
In this paper, we proposed GRExplainer, a universal, efficient, and user-friendly method for explaining TGNNs. GRExplainer extracts node sequences as a unified feature representation, making it applicable to both snapshot-based and event-based TGNNs. By integrating BFS and temporal information, it optimizes node sequence extraction and significantly enhances explanation efficiency. Additionally, GRExplainer employs an RNN-based generation model to produce coherent and connected explanations with minimal human innervation, ensuring user-friendliness. Experimental results across six real-world datasets demonstrated that GRExplainer surpasses the state-of-the-art explanation methods in explaining both TGNNs types in terms of accuracy, efficiency, and user-friendliness.

\bibliographystyle{unsrt}
\bibliography{reference}

% \section*{References}
% References follow the acknowledgments in the camera-ready paper. Use unnumbered first-level heading for
% the references. Any choice of citation style is acceptable as long as you are
% consistent. It is permissible to reduce the font size to \verb+small+ (9 point)
% when listing the references.
% Note that the Reference section does not count towards the page limit.
% \medskip

% {
% \small

% [1] Alexander, J.A.\ \& Mozer, M.C.\ (1995) Template-based algorithms for
% connectionist rule extraction. In G.\ Tesauro, D.S.\ Touretzky and T.K.\ Leen
% (eds.), {\it Advances in Neural Information Processing Systems 7},
% pp.\ 609--616. Cambridge, MA: MIT Press.

% [2] Bower, J.M.\ \& Beeman, D.\ (1995) {\it The Book of GENESIS: Exploring
%   Realistic Neural Models with the GEneral NEural SImulation System.}  New York:
% TELOS/Springer--Verlag.

% [3] Hasselmo, M.E., Schnell, E.\ \& Barkai, E.\ (1995) Dynamics of learning and
% recall at excitatory recurrent synapses and cholinergic modulation in rat
% hippocampal region CA3. {\it Journal of Neuroscience} {\bf 15}(7):5249-5262.
% }

%%%%%%%%%%%%%%%%%%%%%%%%%%%%%%%%%%%%%%%%%%%%%%%%%%%%%%%%%%%%
\newpage
\appendix

% \section{Technical Appendices and Supplementary Material}
% Technical appendices with additional results, figures, graphs and proofs may be submitted with the paper submission before the full submission deadline (see above), or as a separate PDF in the ZIP file below before the supplementary material deadline. There is no page limit for the technical appendices.

\section{Theoretical Proofs} \label{theoretical_proof}
\subsection{Proof of Equation~\ref{problem_formula}} \label{appendix_problem}
$I(Y_f[e]; G_{sub})$ can be approximated by the cross-entropy between the model's prediction given the explanation subgraph ${G_\text{sub}}$ and the target prediction ${Y_f[e]}$.

\begin{equation}
\begin{aligned}
\max \, I(Y_f[e]; G_\text{sub}) 
&\Leftrightarrow \min -I(Y_f[e];G_\text{sub}) \\
&= \min \big[ H(Y_f[e] \mid G_\text{sub}) - H(Y_f[e]) \big] \\
&\Leftrightarrow \min H(Y_f[e] \mid G_\text{sub}) \\
&= \min -\sum_{c\in\mathcal{C}} \mathbbm{1}(Y_f[e]=c) \log(f(G_\text{sub})[e])
\end{aligned}
\label{eq:mutual_info_decomposition}
\end{equation}
where ${H(\cdot)}$ denotes the entropy function and ${H(Y_f[e])}$ remains constant during the explanation phase.

\subsection{Proof of the BFS Property} \label{appendix_bfs}
We use the following observation to prove the BFS property:

\textbf{Observation}. \textit{In a BFS ordering, if $i < k$, then the children if $v_i$ appear before the children of $v_k$ that are not connected to any $v_{i'}$ with $1 \le i' \le i$.}

By definition of BFS, all neighbors of a node $v_i$ include its parent of in the BFS tree, , its children (which appear consecutively in the ordering), and possibly some children of earlier nodes $v_{i'}$ for $1 \le i' \le i$ that are also connected to $v_i$. Therefore, if $(v_i, v_{j-1}) \in E$ but $(v_i, v_j) \not \in E$, then $v_{j-1}$ is the last children of $v_i$ in the BFS ordering. It follows that $(v_{i}, v_{j'}) \not \in E$, $\forall j \le j' \le n$.

Now consider any \( i' \in [i] \), and suppose \( (v_{i'}, v_{j'-1}) \in E \) but \( (v_{i'}, v_{j'}) \notin E \). By the \textbf{Observation}, we have \( j' < j \). From the conclusion above, we then have $(v_{i'}, v_{j''}) \notin E$, $\forall j' \le j'' \le n$.
In particular, $(v_{i'}, v_{j''}) \notin E$, $\forall j \le j'' \le n$. This holds for all \( i' \in [i] \), and thus we prove that $(v_{i'}, v_{j'}) \notin E$, $\forall\ 1 \le i' \le i,\ j \le j' \le n$.

\section{Proposed Approach} \label{algorithm}
\subsection{Algorithm of Explanation on Snapshot-based TGNNs} \label{algorithm_snapshot}
Algorithm~\ref{procedure_snapshot} presents the core steps of GRExplainer for generating an explanation subgraph over a sequence of snapshots $\{G_1, G_2, \dots, G_T\}$. For each snapshot, a local explanation is obtained by optimizing a generation model guided by the loss function. The final explanation $G_{\text{sub}}$ is constructed by aggregating subgraphs across all snapshots.

\begin{algorithm}[!htbp]
\footnotesize
\SetCommentSty{small}
\LinesNumbered
\caption{GRExplainer Explanation on Snapshot-based TGNNs}
\label{procedure_snapshot}

\KwIn{An edge to be explained ${e}$, a sequence of snapshot graphs $\{G_{1}, G_{2}, \dots, G_{T}\}$, the target model ${f}$, original prediction ${y}$;}
\KwOut{Final explanation subgraph $G_{\text{sub}}$;}

\Comment{\textbf{Initialization}}

Initialize the final explanation subgraph: $G_{\text{sub}} \leftarrow \emptyset$;

\For{each snapshot $G_{t}$ in $\{G_{1}, G_{2}, \dots, G_{T}\}$}{

    \Comment{\textbf{Preparation}}

    Obtain the neighboring nodes ${V}_{\text{subgraph}}$;

    Construct the neighboring subgraph adjacency matrix $A_{\text{subgraph}}$;

    Construct the BFS sequence ${V}_{\text{sequence}}$;

    Obtain the retained  matrix $A_{\text{retained}}$ using Eq.~\ref{A_{retained}};

    \Comment{\textbf{Generation Model Optimization}}

    \For{${epoch}$ = 1 to ${num}_{epoch}$}{
        Zero gradients for the generation model;
        
        \Comment{\textbf{Explanation Subgraph Generation}}
        
        Compute the hidden state ${h_{i,1}^1}$ of the edge-level RNN ${{f}_\text{output}}$ using the graph-level RNN ${{f}_\text{rnn}}$ with Eq.~\ref{rnn};

        Compute the adjacency vector ${S_i}$ using the edge-level RNN with Eq.~\ref{output};

        Obtain the edge existence probability vector $P_{\text{edge}}^{(i)}$ through MLP using Eq.~\ref{edge_p};

        Construct the explanation subgraph $G_{\text{sub}}^t$ based on ${P}_{\text{edge}}$;

        Obtain the prediction result $\hat{y}$ by feeding $G_{\text{sub}}^t$ into the target model $f$;

        Compute the loss using Eq.~\ref{loss};

        Update parameters of the generation model;
    }

    Add $G_{\text{sub}}^t$ to the final explanation subgraph: $G_{\text{sub}} \leftarrow G_{\text{sub}} \cup G_{\text{sub}}^t$;
}

\textbf{return} $G_{\text{sub}}$

\end{algorithm}

\subsection{Algorithm of Explanation on Event-based TGNNs} \label{algorithm_event}
Algorithm~\ref{procedure_event} presents the core steps of GRExplainer for generating an explanation subgraph in event-based TGNNs. A sub-event group is first extracted and ordered by timestamps. The generation model is then optimized under the same loss guidance. The resulting explanation $G_{\text{sub}}$ reflects the model's decision-making process over temporal event sequences.

\begin{algorithm}[!htbp]
\footnotesize
\SetCommentSty{small}
\LinesNumbered
\caption{GRExplainer Explanation on Event-based TGNNs}
\label{procedure_event}

\KwIn{An edge to be explained ${e}$, the target model ${f}$, original prediction ${y}$;
}
\KwOut{Explanation subgraph ${{G}_\text{sub}}$;}

\Comment{\textbf{Preparation}}

Obtain the sub-event group ${e}_\text{subgraph}$;

Construct the node sequence by timestamps ;

Obtain the retained matrix ${{A}_\text{retained}}$ using Eq.~\ref{A_{retained}};

\Comment{\textbf{Generation Model Optimization}}

\For{${epoch}$ = 1 to ${num}_{epoch}$ do}
    {
        Zero gradients for the generation model;
        
        \Comment{\textbf{Explanation Subgraph Generation}}
        
        Compute the hidden state ${h_{i,1}^1}$ of the edge-level RNN ${{f}_\text{output}}$ using the graph-level RNN ${{f}_\text{rnn}}$ with Eq.~\ref{rnn};

        Compute the adjacency vector ${S_i^\pi}$ using the edge-level RNN with Eq.~\ref{output};
        
        Obtain the edge existence probability vector $P_{\text{edge}}^{(i)}$ through MLP using Eq.~\ref{edge_p};

        Construct the generated subgraph ${{G}_\text{sub}}$ based on ${{P}_\text{edge}}$;
        
        Obtain the prediction result $\hat{y}$ by feeding ${G}_\text{sub}$ into the target model ${f}$;

        Compute the ${loss}$ using Eq.~\ref{loss};

        Update parameters of the generation model;
    }

\textbf{return} $G_{\text{sub}}$

\end{algorithm}

\section{Experiments} \label{appendix_experiments}
\subsection{Datasets} \label{appendix_dataset}
We adopt six real-world datasets from different domains, with half consisting of snapshot graphs and the other half consisting of event graphs. Detailed information about these datasets is introduced below and provided in Tables~\ref{snapshot_dataset} and~\ref{event_dataset}.
\begin{itemize}
    \item Reddit-Binary~\cite{yanardag2015deep} contains user interaction graphs from two types of subreddits: Q\&A-based and discussion-based. Each graph corresponds to a subreddit over 29 months, where nodes represent users and edges indicate reply relationships. 
    \item Bitcoin-Alpha~\cite{lepomaki2021retaliation} and Bitcoin-OTC~\cite{lepomaki2021retaliation} are who-trusts-whom networks of bitcoin users trading on the bitcoin-otc platform and the btc-alpha platform, respectively. Each of them is a weighted, directed, signed graph, where nodes represent users and edges represent trust ratings ranging from -10 (distrust) to +10 (trust) with timestamps.
    \item Reddit~\cite{kumar2019predicting} is a bipartite interaction network capturing one month of user posts on subreddits, with users and subreddits as nodes and timestamped posting requests as edges.
    \item Wikipedia~\cite{kumar2019predicting} is a bipartite interaction network capturing user edits on Wikipedia pages over a month, with nodes representing users and pages, and edges representing edit behaviors.
    \item Mooc~\cite{kumar2019predicting} is an online education platform network, with nodes representing students and course resources, and edges representing timestamped learning actions (e.g., video viewing).
\end{itemize}

\textit{Data Preparation.} 
For the event graph datasets Reddit, Wikipedia, and Mooc, we randomly select 100 time indices within the temporal range of the validation set. The corresponding temporal edges are used as the predictions to be explained. For the Reddit-Binary dataset, we randomly select 100 edges from a single snapshot as the predictions to be explained. Due to the sparsity of the Bitcoin-Alpha and Bitcoin-OTC datasets, we impose a specific filtering criterion: the selected edges must have at least 8 edges in the subgraph formed by their two-hop neighbors. Based on this criterion, we select 11 edges from Bitcoin-Alpha and 47 edges from Bitcoin-OTC for explanation.

\begin{table}[htb]
    \centering
    \small
    \caption{Statistics of snapshot graph datasets: Reddit-Binary, Bitcoin-Alpha, and Bitcoin-OTC.}
    \label{snapshot_dataset}
    \vspace{0.5em}
    \begin{tabular}{lcccc}
        \toprule
        Dataset & Domain & \#Nodes & \#Edges & Edge Weights \\
        \midrule
        Reddit-Binary & Social    & 55,863  & 858,490  & -1 or +1 \\
        Bitcoin-Alpha & Financial & 3,783   & 24,186   & -10 to +10 \\
        Bitcoin-OTC   & Financial & 5,881   & 35,592   & -10 to +10 \\
        \bottomrule
    \end{tabular}
\end{table}

\begin{table}[htb]
    \centering
    \small
    \caption{Statistics of event graph datasets: Reddit, Wikipedia, and Mooc.}
    \label{event_dataset}
    \vspace{0.5em}
    \begin{tabular}{lccccc}
        \toprule
        Dataset & Domain & \#Nodes (Src/Dst) & \#Edges & \#Node/Edge Features & Time Span \\
        \midrule
        Reddit     & Social     & 10,984 (10,000/984) & 672,447 & 0/172 & 1 month  \\
        Wikipedia  & Social & 9,227 (8,227/1,000) & 157,474 & 0/172 & 1 month  \\
        Mooc       & Education  & 7,144 (7,047/97)    & 411,749 & 0/4   & 29 months \\
        \bottomrule
    \end{tabular}
\end{table}

\subsection{Target Models}
We adopt three popular state-of-the-art TGNNs, including one snapshot-based (EvolveGCN~\cite{pareja2020evolvegcn}) and two event-based models (TGAT~\cite{xu2020inductive} and TGN~\cite{rossi2020temporal}), as target models to be explained. EvolveGCN~\footnote{\url{https://github.com/IBM/EvolveGCN}} updates the weight matrix of GCN models across different snapshots using RNNs. TGAT~\footnote{\url{https://github.com/StatsDLMathsRecomSys/Inductive-representation-learning-on-temporal-graphs}} introduces a temporal attention mechanism that aggregates a node’s historical neighbors in chronological order, along with a time encoding function to capture fine-grained temporal patterns. TGN~\footnote{\url{https://github.com/twitter-research/tgn}} extends this approach by incorporating a memory module that continuously maintains and updates each node’s temporal state throughout the evolution of the graph.

\subsection{Baseline Methods}
For snapshot-based TGNN explanation, we choose \textbf{GNNExplainer} as the baseline. GNNExplainer~\footnote{\url{https://github.com/RexYing/gnn-model-explainer}} is the first explanation method for static GNNs, which constructs explanation subgraphs by learning soft masks over edges and node features.

For event-based TGNN explanation, we consider the following strong baselines:
\begin{itemize}
    \item \textbf{T-GNNExplainer}~\footnote{\url{https://github.com/cisaic/tgnnexplainer}} is the first method designed to explain event-based TGNNs. It integrates a navigator with Monte Carlo Tree Search, where the navigator guides the sampling process of the tree search, and the sampled event groups are used to construct explanation subgraphs.
    \item \textbf{TempME}~\footnote{\url{https://github.com/Graph-and-Geometric-Learning/TempME}} is the first generative method for explaining event-based TGNNs. It identifies temporal motifs, computes their importance scores, and constructs explanation subgraphs by selecting motifs with high scores.
\end{itemize}

\subsection{Evaluation Metrics}
\textbf{Fidelity+ (FID+)} is calculated as:
\begin{equation}
{\text{FID+}} = \frac{1}{N}\mathop \sum \limits_{i = 1}^N (f({G^i_{\text{subgraph}}})_{y_i}-f({(\overline{G}^i_{\text{sub}})}_{y_i}),
\label{fidelity+}
\end{equation}
where $N$ denotes the number of explanation instances, $f$ the target TGNN model, and ${y_i}$ the label of the $i$-th instance. $G_\text{subgraph}^i$ represents the computation subgraph of instance $i$, while ${\overline{G}^i_\text{sub}}$ refers to the remaining subgraph obtained by removing the explanation $G_\text{sub}^i$ from $G_\text{subgraph}^i$. A higher FID+ value indicates that the explanation captures more important information for the model’s prediction.

\textbf{The Area Under the Fidelity-Sparsity Curve (AUFSC)} is calculated as:
\begin{equation} 
\text{AUFSC} = \int_0^\lambda \text{FID+}(\lambda) d\lambda, 
\end{equation}
where $\lambda$ is the sparsity level, $\text{FID+}(\lambda)$ is the corresponding fidelity value. The upper limit $\lambda \leq 1$ defines the maximum level of sparsity considered in the evaluation.

\textbf{Cohesiveness} is calculated as:
\begin{equation}
\text{Cohesiveness} = \log\left(\frac{1}{|G_{\text{sub}}| \times (2 - |G_{\text{sub}}|)} \sum_{e_i, e_j \in G_{\text{sub}}, e_i \neq e_j} \cos\left( \frac{|t_i - t_j|}{\Delta T} \right) \times \mathbbm{1}(e_i \sim e_j)\right),
\label{eq:log-cohensive}
\end{equation}
where $|G_{\text{sub}}|$ denotes the number of interactions in the explanation subgraph ${G_{\text{sub}}}$. ${t_i}$ and ${t_j}$ are the timestamps of interactions ${e_i}$ and ${e_j}$, and ${\Delta T}$ is the gap between the maximum and minimum timestamps. The indicator function $\mathbbm{1}(e_i \sim e_j)$ equals 1 if ${e_i}$ and ${e_j}$ are spatially adjacent, i.e., they share a same node. The cosine term measures temporal closeness, assigning higher values to interactions that occur closer in time. Thus, a higher cohesiveness score indicates stronger spatial and temporal correlations between interactions, reflecting better connectivity in the explanation.

\subsection{Implementation Details} \label{appendix_detail}
The explanation generation model was implemented based on the RNN architecture of GraphRNN~\cite{you2018graphrnn}, with the following hyperparameter specifications: The graph-level RNN $f_{\text{rnn}}$ receives an input vector of dimension $M$, which encodes retained graph information, and uses an embedding dimension of 64 and a hidden dimension of 128. The edge-level RNN ($f_{\text{output}}$) has an input dimension of 1, embedding dimension of 64, hidden dimension of 128, a 2-layer GRU structure, and a 16-dimensional hidden output space. The generation model was optimized using the Adam optimizer, with the learning rate of 0.06 determined through grid search. This configuration was applied consistently across all comparative experiments.

\subsection{Ablation Study} \label{appendix_ablation}
To validate the necessity of the sparsity regularization term used in Eq.~\ref{loss}, we construct a variant, named ``w/o Sparsity'', by removing the sparsity constraint on the generated explanations.

% \begin{figure}
\begin{wrapfigure}{r}{0.48\textwidth}
        %\vspace{-0.5cm}
        \centering
        \includegraphics[width=\linewidth]{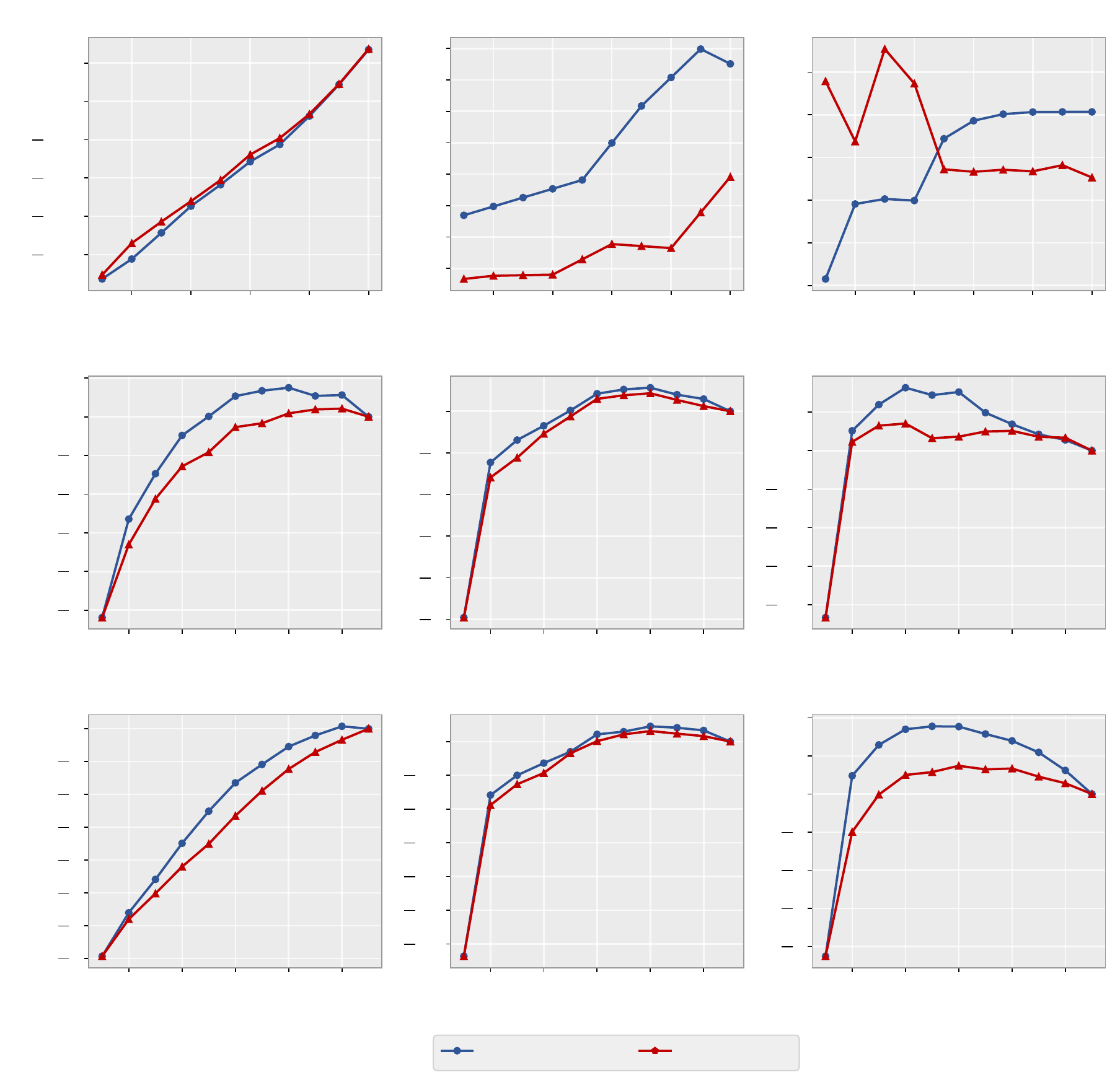}
        %\vspace{-0.5cm}
        \caption{Fidelity-Sparsity comparison with and without the sparsity regularization term.}
       \label{nospa}
       %\vspace{-0.4cm}
% \end{figure}
\end{wrapfigure}
Fig.~\ref{nospa} shows that incorporating the sparsity regularization term consistently improves explanation fidelity for event-based TGNN models (TGAT and TGN) across all sparsity levels. This indicates that the regularization term effectively guides the explanation process toward concise and informative substructures, reducing noise and enhancing accuracy. In contrast, its impact on snapshot-based models is less consistent. On the Bitcoin Alpha dataset, sparsity regularization substantially improves fidelity. However, on Reddit-Binary, it has minimal effect, and on Bitcoin OTC, explanations without regularization outperform those with it at sparsity levels between 0.07 and 0.28. This may be attributed to the model's greater flexibility in retaining high-scoring edges when the regularization is removed. Nonetheless, the absence of regularization introduces randomness, resulting in unstable fidelity due to the GNN’s sensitivity to structural changes.

\subsection{Hyperparameter Analysis} \label{appendix_hyparameters}
\begin{figure}[htbp]
    \centering
    \begin{minipage}[t][][b]{0.48\textwidth} % 图片顶对齐，caption底对齐
        \vspace{0pt} % 基准线重置
        \centering
        \includegraphics[width=\linewidth, height=6cm]{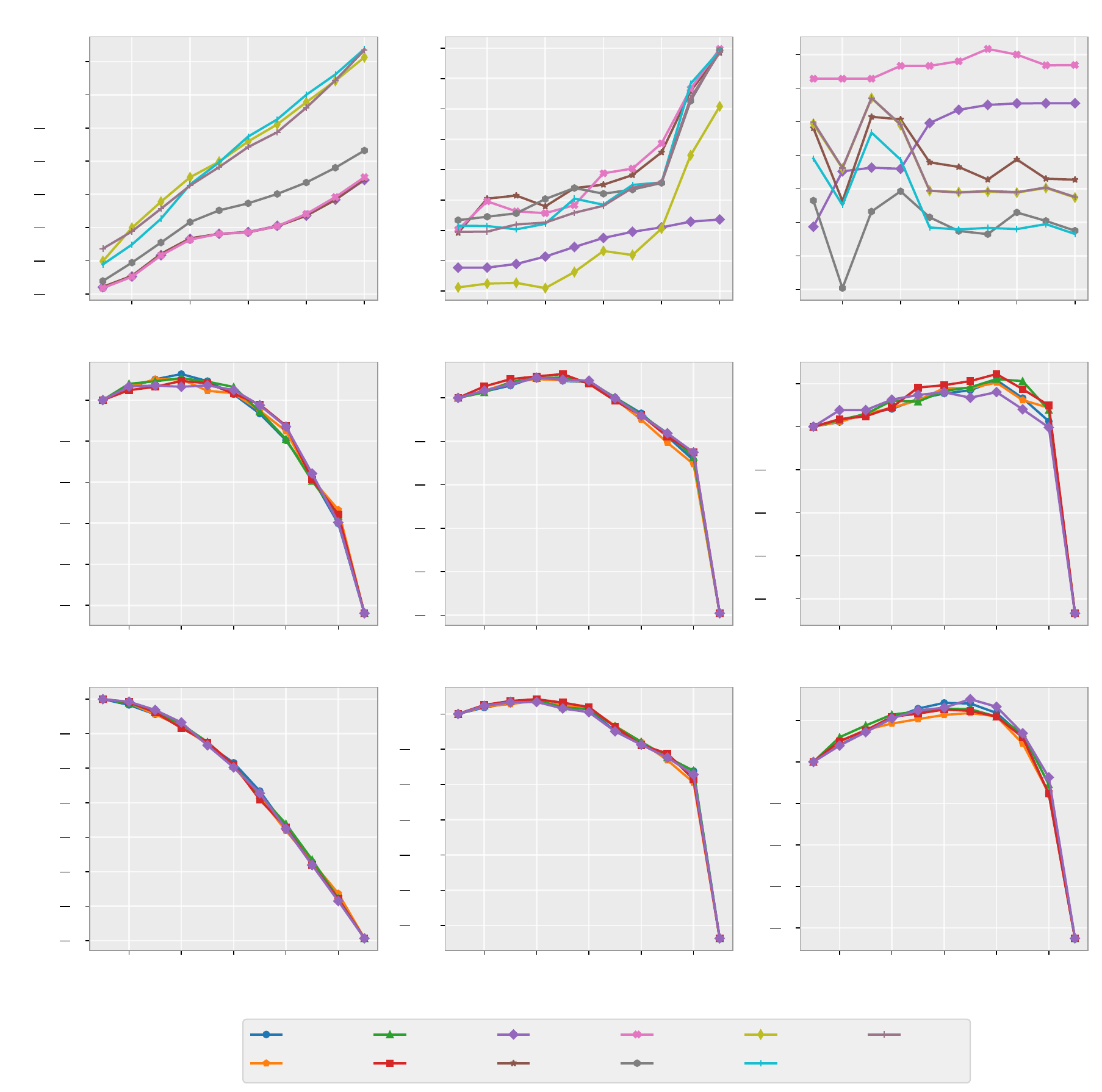} % 推荐固定高度
        \caption{The impact of $\lambda_{\text{size}}$ on fidelity.}
        \label{size}
    \end{minipage}
    \hfill
    \begin{minipage}[t][][b]{0.48\textwidth}
        \vspace{0pt}
        \centering
        \includegraphics[width=\linewidth, height=6cm]{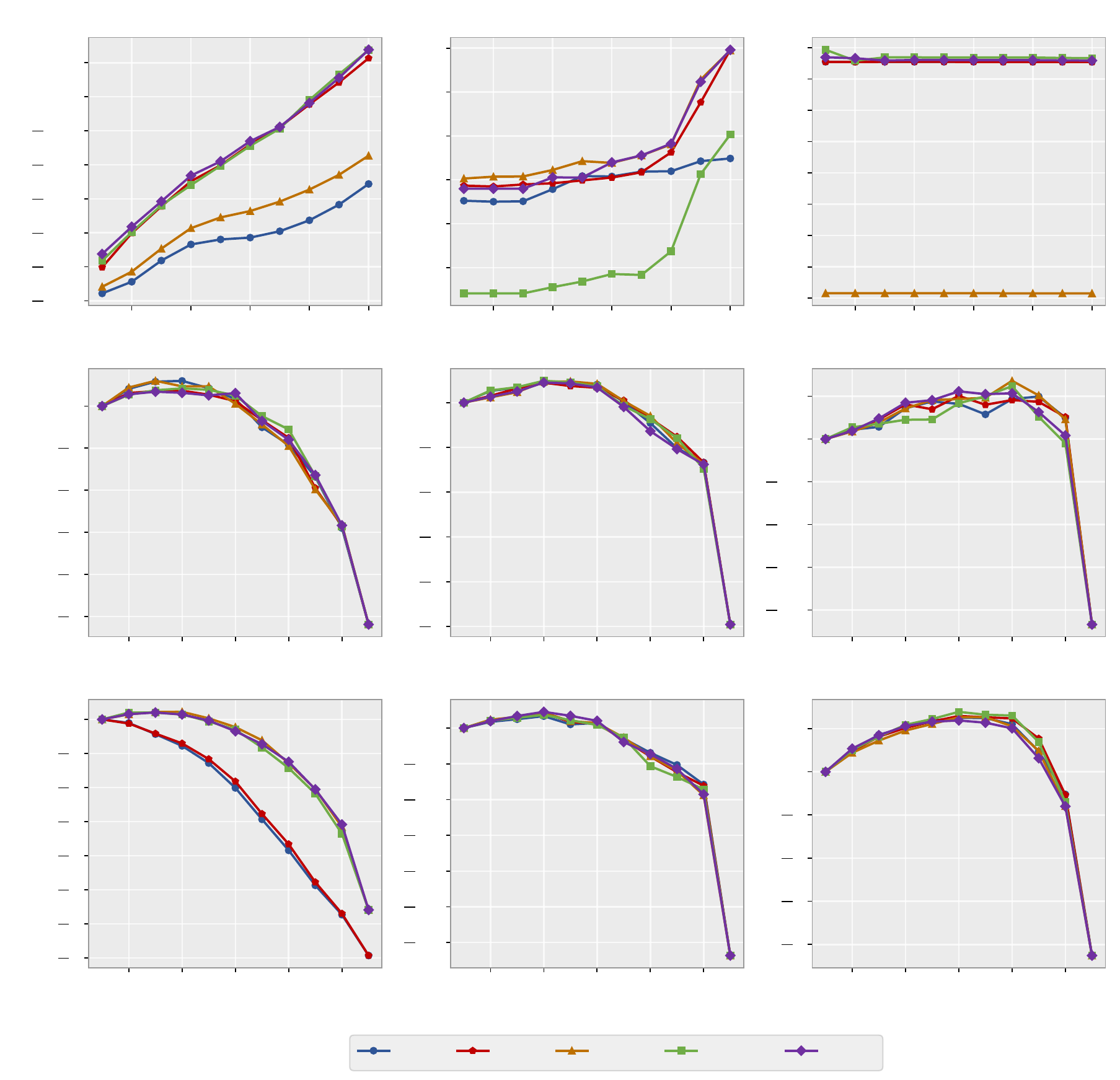}
        \caption{The impact of $\lambda_{\text{weight}}$ on fidelity.}
        \label{weight}
    \end{minipage}
\end{figure}

Table~\ref{param} summarizes the search ranges of $\lambda_{\text{size}}$ and $\lambda_{\text{weight}}$ used in Eq.~\ref{loss}. Their effects on explanation accuracy are illustrated in Fig.~\ref{size} and~\ref{weight}. As shown, both hyperparameters exhibit distinct impacts across different models and datasets.

\begin{table}[ht]
    \centering
    \small
    \caption{Search ranges of $\lambda_{\text{size}}$ and $\lambda_{\text{weight}}$.}
    \vspace{0.5em}
    \label{param}
    \begin{tabular*}{\linewidth}{@{\extracolsep{\fill}}ccc}
         \toprule
        Dataset & $\lambda_{\text{size}}$ & $\lambda_{\text{weight}}$ \\
        \midrule
        Reddit & \{$5{\times}10^{-4}$,$5{\times}10^{-3}$,$5{\times}10^{-2}$,$5{\times}10^{-1}$,$5$\} & \{0,1,10,100,1000\} \\
        Wikipedia & \{$5{\times}10^{-4}$,$5{\times}10^{-3}$,$5{\times}10^{-2}$,$5{\times}10^{-1}$,$5$\} & \{0,1,10,100,1000\} \\ 
        Mooc & \{$5{\times}10^{-4}$,$5{\times}10^{-3}$,$5{\times}10^{-2}$,$5{\times}10^{-1}$,$5$\} & \{0,1,10,100,1000\} \\
        Reddit-Binary & \{$5{\times}10^{-10}$,$5{\times}10^{-9}$,$5{\times}10^{-8}$,$5{\times}10^{-7}$,$5{\times}10^{-6}$,$5{\times}10^{-5}$,$5{\times}10^{-4}$\} & \{0,1,10,100,1000\} \\
        Bitcoin-Alpha & \{$5{\times}10^{-10}$,$5{\times}10^{-9}$,$5{\times}10^{-8}$,$5{\times}10^{-7}$,$5{\times}10^{-6}$,$5{\times}10^{-5}$,$5{\times}10^{-4}$\} & \{0,1,10,100,1000\} \\
        Bitcoin-OTC & \{$5{\times}10^{-10}$,$5{\times}10^{-9}$,$5{\times}10^{-8}$,$5{\times}10^{-7}$,$5{\times}10^{-6}$,$5{\times}10^{-5}$,$5{\times}10^{-4}$\} & \{0,1,10,100,1000\} \\
        \bottomrule
    \end{tabular*}
\end{table}

For $\lambda_{\text{size}}$, TGAT consistently achieves optimal explanation fidelity with $\lambda_{\text{size}}$ around $5 \times 10^{-3}$ across Wikipedia, Mooc, and Reddit datasets. In contrast, TGN exhibits more varied sensitivity, requiring $\lambda_{\text{size}}$ values of approximately $5 \times 10^{-4}$ for Mooc, $5 \times 10^{-3}$ for Wikipedia, and $5 \times 10^{-2}$ for Reddit to maintain optimal performance. Additionally, EvolveGCN shows a clear scaling trend, with smaller datasets like Bitcoin-Alpha, Bitcoin-OTC, and Reddit-Binary achieving optimal performance at smaller $\lambda_{\text{size}}$ values ranging from $5 \times 10^{-5}$ to $5 \times 10^{-8}$. This inverse relationship between dataset scale and regularization strength highlights fundamental differences in how each model processes structural information. TGAT displays stable behavior across datasets, while TGN and EvolveGCN require dataset-specific tuning.

For $\lambda_{\text{weight}}$, TGAT performs best with values around 10 on Mooc, 100 on Reddit, and shows low sensitivity (1–100) on Wikipedia. In contrast, TGN requires a higher value (approximately 1000) for optimal fidelity on Wikipedia, whereas EvolveGCN is largely insensitive to $\lambda_{\text{weight}}$ on Bitcoin-OTC but performs best around 10 on Bitcoin-Alpha. These results highlight the model-specific strategies employed by different architectures in incorporating $\lambda_{\text{weight}}$ during explanation generation.

Overall, we have two observations: (i) Larger datasets tend to require higher $\lambda_{\text{size}}$ values. (2) Model architectures exhibit distinct sensitivity patterns to hyperparameter changes. Based on these observations, we recommend the following initialization strategies: For large-scale graphs, set $\lambda_{\text{size}} \approx 5 \times 10^{-3}$ and $\lambda_{\text{weight}} \approx 100$; For smaller graphs, begin with $\lambda_{\text{size}} \approx 5 \times 10^{-4}$ and $\lambda_{\text{weight}} \approx 10$, followed by model-specific tuning.

\subsection{Visualization of Explanations} \label{appendix_visualization}
\begin{wrapfigure}{r}{0.58\textwidth}
    \vspace{-0.5cm}
    \subfigure[GRExplainer]
    {\includegraphics[width=0.19\textwidth]{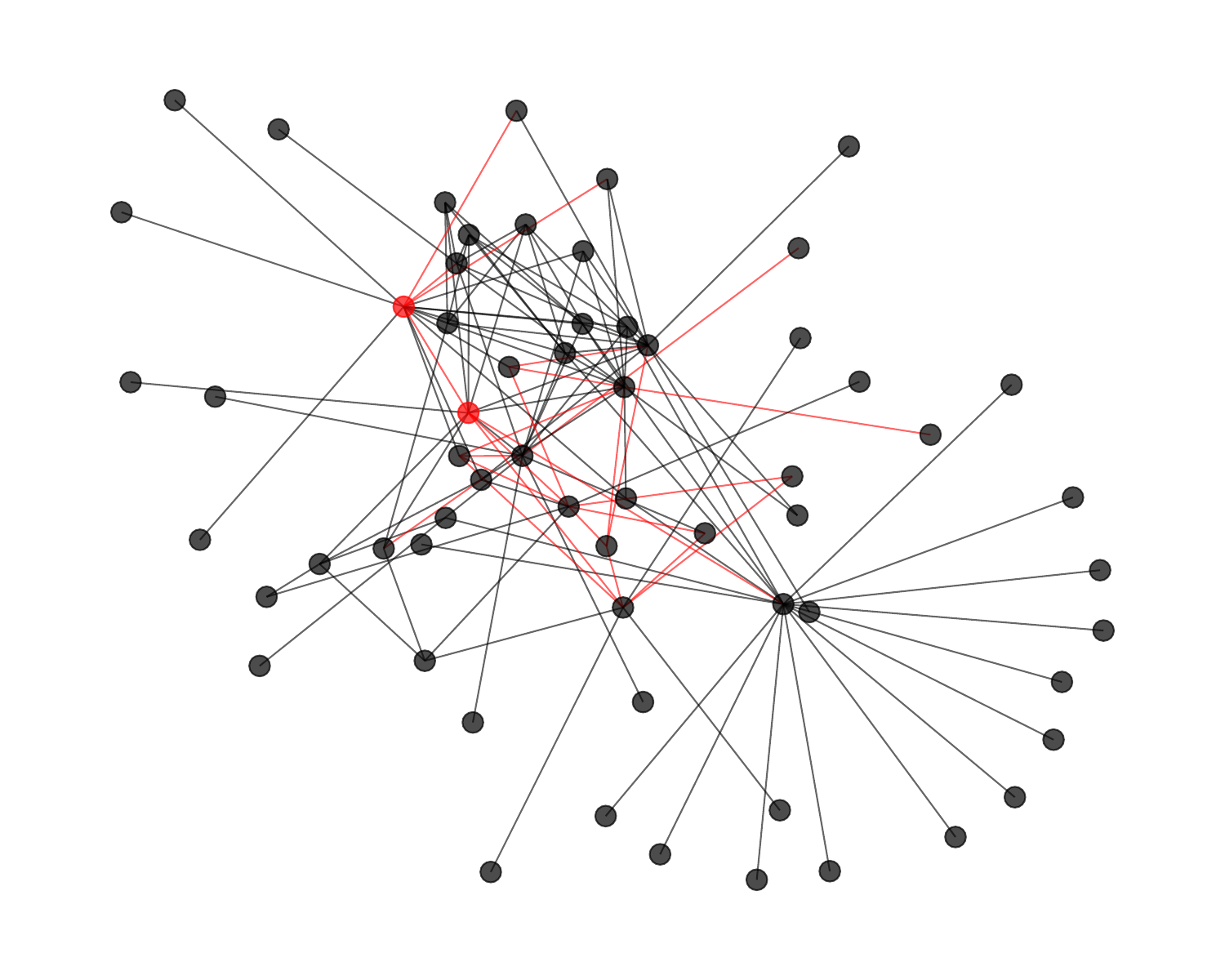}}
    \subfigure[T-GNNExplainer]
    %\subfigure[\footnotesize T-GNNExp]
    {\includegraphics[width=0.19\textwidth]{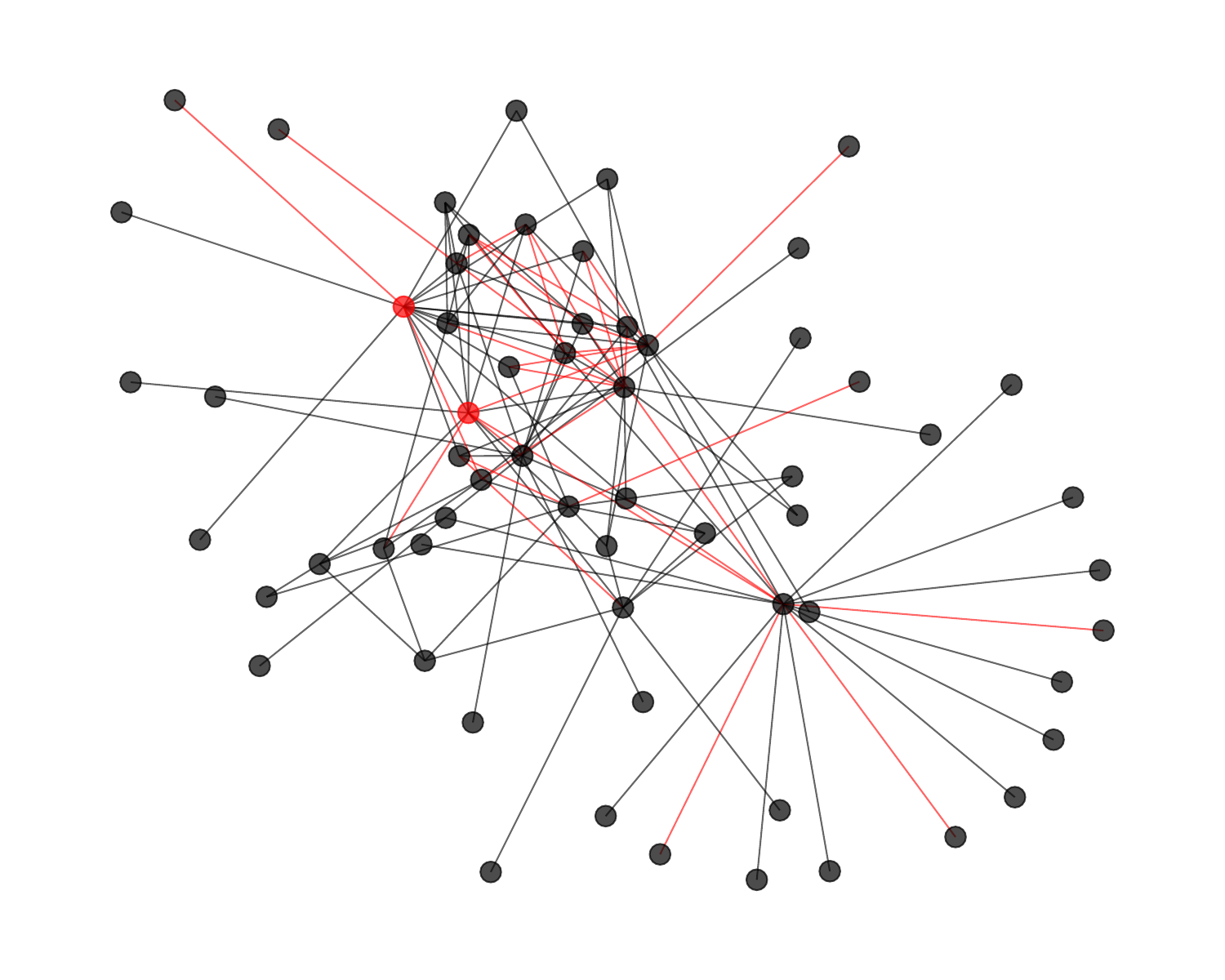}}
    \subfigure[TempME]  % 缩放90%
    %\subfigure[\footnotesize TempME]
    {\includegraphics[width=0.19\textwidth]{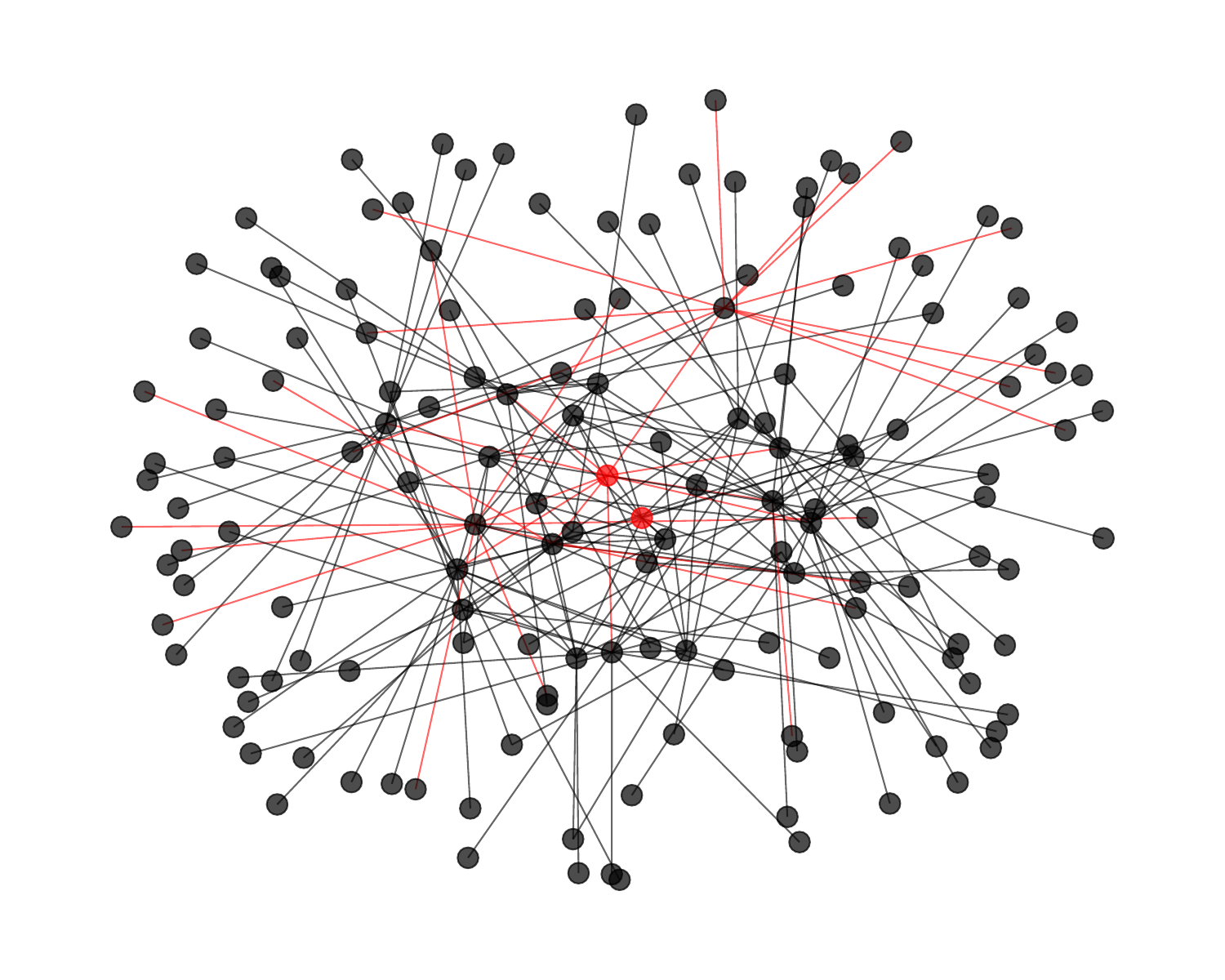}}
    \vspace{-2mm}
    \caption{Visualization of explanation subgraphs.}
    \label{explanation_graph}
\end{wrapfigure}
To further validate the user-friendliness of GRExplainer, we visualize the explanation subgraphs for TGAT predictions on the Mooc dataset, generated by GRExplainer, T-GNNExplainer, and TempME. As shown in Fig.~\ref{explanation_graph}, the red nodes represent the source and target nodes of the edge to be explained, while the red edges denote the explanation subgraphs produced by each method. The following observations can be made: (i) GRExplainer generates explanation subgraphs that are spatially concentrated around the local neighborhood of the target edge, thereby enhancing interpretability. (ii) GRExplainer uniquely retains the historical interaction patterns between the source and target nodes—an essential capability lacking in the baseline methods. These temporal patterns are critical for understanding the decision-making process of temporal GNNs in link prediction tasks. In summary, GRExplainer demonstrates superior user-friendliness by producing cohesive explanations without requiring prior knowledge, making it a practical and accessible choice for real-world applications.

\section{Discussion}
\subsection{Broader Impacts} \label{broader_impacts}
GRExplainer holds significant potential across various domains, including finance, social networks and cybersecurity, where understanding dynamic relationships is critical. In finance, it can help auditors and compliance officers decipher suspicious transaction patterns over time. In social networks, it can reveal the evolution of interactions and the propagation of information or influence, contributing to community and misinformation analysis. In cybersecurity, it interprets alerts related to anomalous network activity and supports threat detection and response. Furthermore, its universal design makes it feasible to be applied into both snapshot-based and event-based TGNNs, its efficiency enables scalable real-time analysis, and its user-friendliness provides intuitive explanations for both experts and normal people. These benefits make GRExplainer a valuable tool for deploying transparent and trustworthy AI in high-risk and time-sensitive environments.

\subsection{Limitations and Future Work} \label{limitation}
GRExplainer operates at the instance level (i.e., explaining one instance at a time), which limits its efficiency when applied into a large number of graphs. Future work could focus on developing class-level explanation methods capable of explaining groups of instances with similar characteristics~\cite{li2025can}, thereby enhancing efficiency. Moreover, in graph classification tasks, GRExplainer needs to consider all nodes within the entire graph, leading to high computational overhead. To address this, future directions include exploring more efficient explanation strategies, such as node selection or structural summarization techniques. Finally, GRExplainer is primarily designed for explaining predictions made by homogeneous TGNNs and is not directly applicable to heterogeneous TGNNs. Extending it to heterogeneous graphs remains an important direction, which may require type-aware mechanisms and relation-specific modeling to accommodate diverse graph structures.

%%%%%%%%%%%%%%%%%%%%%%%%%%%%%%%%%%%%%%%%%%%%%%%%%%%%%%%%%%%%

\newpage

\end{document}